%% file: main.tex
\documentclass[10pt,twocolumn,letterpaper]{article}

\usepackage[pagenumbers]{cvpr}
\usepackage{colortbl}
\usepackage{multirow}
\usepackage{algorithm}
\usepackage{algpseudocode}
\usepackage{bm}

\input{preamble}

\definecolor{cvprblue}{rgb}{0.21,0.49,0.74}
\usepackage[pagebackref,breaklinks,colorlinks,allcolors=cvprblue]{hyperref}

\title{Image-Free Timestep Distillation via Continuous-Time Consistency with Trajectory-Sampled Pairs}

\author{
\begin{tabular}{c}
Bao Tang \quad Shuai Zhang \quad Yueting Zhu \quad Jijun Xiang \quad Xin Yang \\
Li Yu \quad Wenyu Liu \quad Xinggang Wang$^{\dagger}$ \\
Huazhong University of Science and Technology
\end{tabular}
}

\begin{document}

\twocolumn[{
	\renewcommand\twocolumn[1][]{#1}
	\maketitle
    \vspace{-25pt}
	\begin{center}
		\centering
        \includegraphics[width=\textwidth]{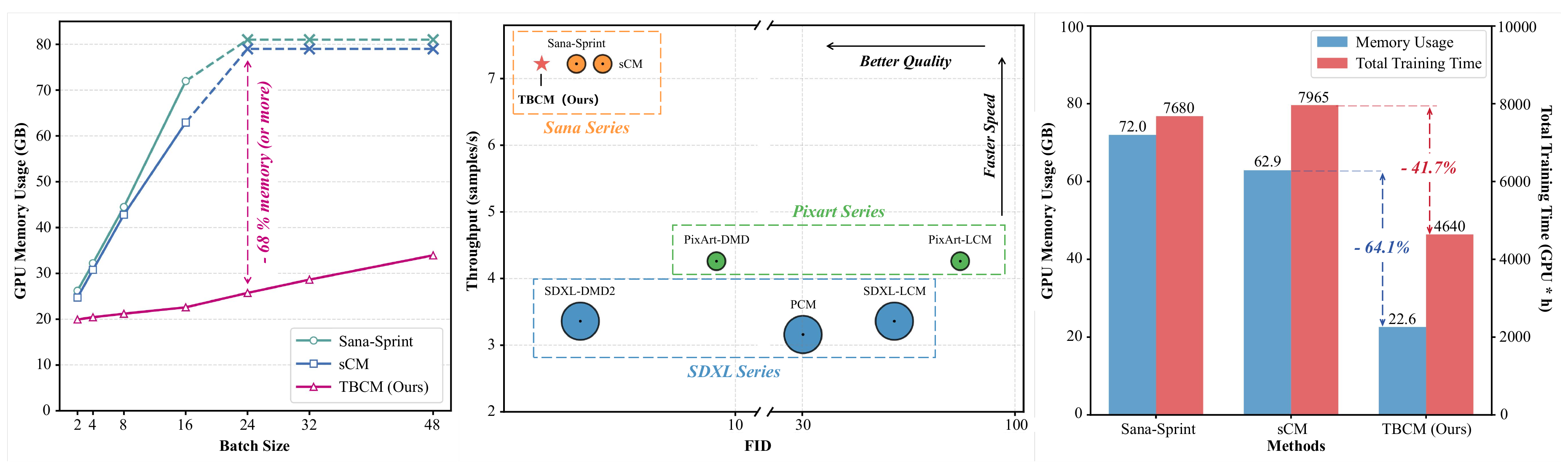}
        \captionof{figure}{\textbf{Comprehensive Comparison.} \textit{\textbf{Left:}} GPU memory usage versus batch size during training, where \textit{Batch Size} denotes the number of samples actually involved in optimization. \textit{\textbf{Middle:}} Comparison of FID scores and throughput across different methods; the marker size indicates the model parameter count. \textit{\textbf{Right:}} GPU memory consumption and total training time under identical training configurations.}
        \label{fig:result}
	\end{center}
	\label{teaserfig}
}]

\begingroup
\makeatletter
\renewcommand\@makefnmark{}
\renewcommand\thefootnote{} 
\makeatother

\footnote{
$^{\dagger}$ Corresponding author (\texttt{xgwang@hust.edu.cn}).
}

\addtocounter{footnote}{-1}
\endgroup

\input{sec/0_Abstract}
\vspace{-1.5em}
\input{sec/1_Intro}
\input{sec/2_Related_Work}
\input{sec/3_Preliminaries}
\input{sec/4_Methods}
\input{sec/5_Experiments}
\input{sec/6_Conclusion}

{
    \small
    \bibliographystyle{ieeenat_fullname}
    \bibliography{main}
}

\input{sec/X_suppl}

\end{document}

%% file: preamble.tex
\input{math.tex}

\newcommand{\boldparagraph}[1]{\vspace{.5em}\noindent\textbf{#1}}


%% file: math.tex
\usepackage{amsmath,amsfonts,bm}

\newcommand{\F}{\mF}

\newcommand{\x}{\vx}
\newcommand{\y}{\vy}
\newcommand{\z}{\vz}

\newcommand{\dm}{\mathrm{d}}

\def\eqref#1{Eq.~(\ref{#1})}

\def\1{\bm{1}}

\def\vzero{{\bm{0}}}

\def\vg{{\bm{g}}}

\def\vx{{\bm{x}}}
\def\vy{{\bm{y}}}
\def\vz{{\bm{z}}}

\def\mF{{\bm{F}}}

\def\mI{{\bm{I}}}

\DeclareMathAlphabet{\mathsfit}{\encodingdefault}{\sfdefault}{m}{sl}
\SetMathAlphabet{\mathsfit}{bold}{\encodingdefault}{\sfdefault}{bx}{n}

\def\gD{{\mathcal{D}}}

\def\gL{{\mathcal{L}}}

\def\gN{{\mathcal{N}}}

\newcommand{\bx}{\bm{x}}
\newcommand{\bz}{\bm{z}}

%% file: sec/0_Abstract.tex
\begin{abstract}

Timestep distillation is an effective approach for improving the generation efficiency of diffusion models. The Consistency Model (CM), as a trajectory-based framework, demonstrates significant potential due to its strong theoretical foundation and high-quality few-step generation.
Nevertheless, current continuous-time consistency distillation methods still rely heavily on training data and computational resources, hindering their deployment in resource-constrained scenarios and limiting their scalability to diverse domains.
To address this issue, we propose \textbf{Trajectory-Backward Consistency Model (TBCM)}, which eliminates the dependence on external training data by extracting latent representations directly from the teacher model's generation trajectory. Unlike conventional methods that require VAE encoding and large-scale datasets, our self-contained distillation paradigm significantly improves both efficiency and simplicity.
Moreover, the trajectory-extracted samples naturally bridge the distribution gap between training and inference, thereby enabling more effective knowledge transfer.
Empirically, TBCM achieves \textbf{6.52 FID} and \textbf{28.08 CLIP} scores on MJHQ-30k under one-step generation, while reducing training time by approximately \textbf{40\%} compared to Sana-Sprint and saving a substantial amount of GPU memory, demonstrating superior efficiency without sacrificing quality.
We further reveal the diffusion-generation space discrepancy in continuous-time consistency distillation and analyze how sampling strategies affect distillation performance, offering insights for future distillation research.
GitHub Link: \url{https://github.com/hustvl/TBCM}.

\end{abstract}

%% file: sec/1_Intro.tex
\section{Introduction}
\label{sec:intro}

\begin{figure*}[th]
    \centering
    \includegraphics[width=1.0\textwidth]{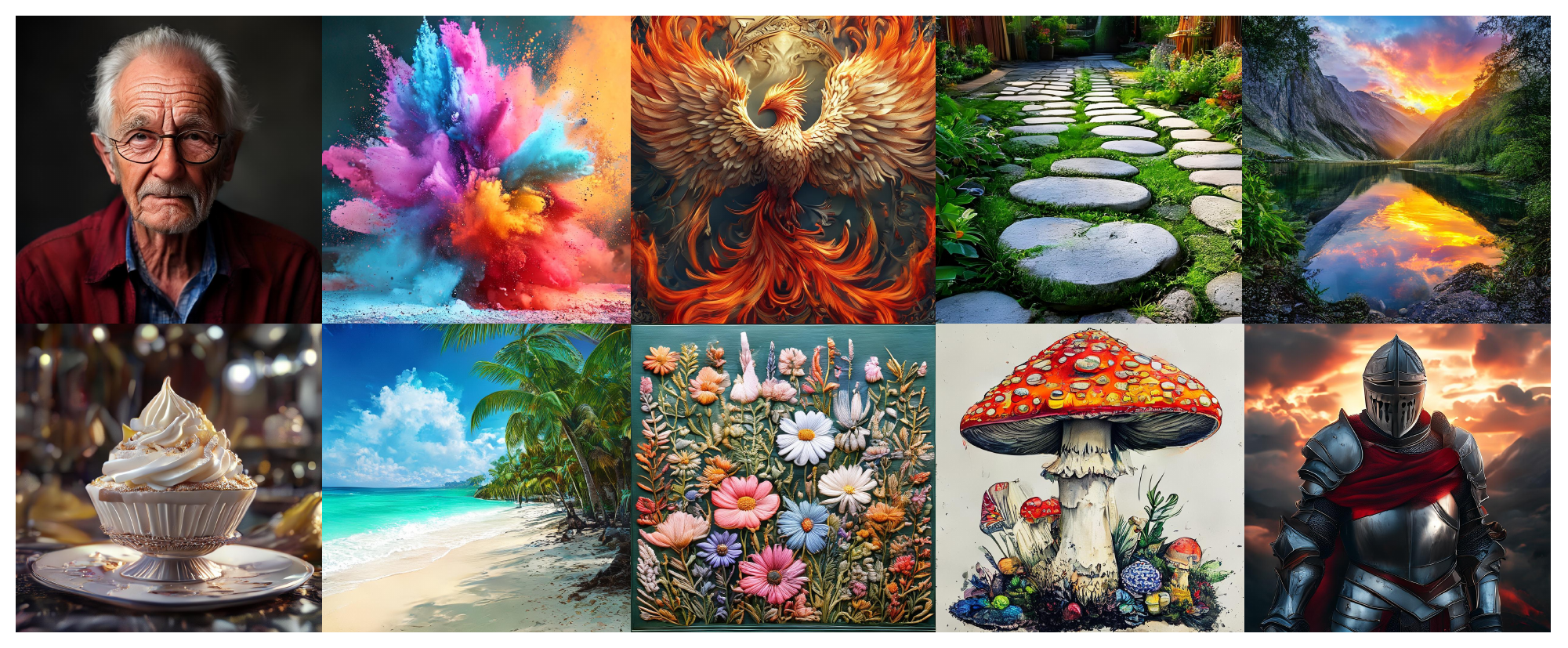}
    \caption{\textbf{One Step Generation Results.} High-resolution (1024×1024) images generated by our one-step generator distilled from the Sana 0.6B model using the proposed TBCM. More results with different sampling steps are provided in the Appendix.}
    \vspace{-1em}
    \label{fig:generate}
\end{figure*}

Diffusion models have achieved remarkable success across a wide range of generative tasks, such as image~\citep{ldm, imagen, dalle, pixart, sd3, flux} and video~\citep{svd, opensora_plan, videocrafter, cogvideox, wan} synthesis. 
However, their generation process typically requires dozens or even hundreds of iterative denoising steps~\citep{ddpm, iddpm, ddim}, leading to extremely long inference times and high computational costs, which severely limit their real-world applicability.
To address this issue, a line of research known as timestep distillation~\citep{direct, progressive, cm, scm, add, ladd, dmd, dmd2} has emerged, which aims to transfer the multi-step diffusion process into a compact student model capable of generating high-quality samples in only a few steps.

Among these efforts, \textbf{Consistency Models (CMs)}~\citep{cm} have recently attracted significant attention for their elegant formulation and training efficiency.
They leverage the consistency constraint, which eliminates the need for supervision from diffusion model samples~\citep{direct, progressive}, thereby avoiding the computational overhead of generating synthetic datasets and circumventing the inherent training instability of adversarial methods.
\textbf{Continuous-Time Consistency Distillation (CTCD)}, as the continuous-time formulation of CMs, removes the discretization error present in discrete-time variants and achieves superior distillation quality.
Furthermore, \textbf{sCM}~\citep{scm} establishes a training paradigm based on the TrigFlow architecture, which effectively stabilizes the training of CTCD and makes it a highly promising approach for timestep distillation.

Although sCM partially addresses the stability issues of CTCD, it still faces several limitations, such as high data requirements, expensive training costs, and limited applicability.
Moreover, during the distillation process, its target sample points are still generated through forward diffusion following the pretraining paradigm of diffusion models. These samples inherently differ from the model’s actual inference trajectory, which constrains its potential to further improve the quality of one-step generation.

In this work, we propose \textbf{TBCM}, an image-free distillation framework that harnesses the teacher model’s generative ability by sampling along its inference trajectories, which mitigates training–inference inconsistency and enables fully latent-space consistency distillation.
By removing VAE involvement and performing multiple trajectory samples per prompt, TBCM significantly lowers GPU memory usage, reduces training time, and is conveniently transferable to other diffusion-based tasks.

In addition, we conduct a detailed investigation into how sampled points from different trajectories influence the training performance, providing strong empirical evidence for the pivotal role of sampling scheme in the consistency distillation process.
To further enhance the distillation quality, we adjust the weighting of the unstable term in the sCM loss, achieving a more balanced optimization objective and improved training results.

Through these efforts, we successfully realize efficient timestep distillation for text-to-image (T2I) models under the image-free setting.
As shown in Fig.~\ref{fig:result}, our approach achieves an outstanding FID of 6.52 and a CLIP score of 28.08 under one-step generation on the MJHQ-30k~\citep{playground} benchmark, while significantly reducing both the training time and GPU memory consumption compared with Sana-Sprint~\citep{sana-sprint} and the standard sCM~\citep{scm} baseline.

Overall, our main contributions can be summarized as follows:

\begin{table*}[t]
    \centering
    \renewcommand{\arraystretch}{1.15}
    \setlength{\tabcolsep}{6pt}
    \small
    \begin{tabular}{ccccccc}
        \toprule
        \textbf{Method} & \textbf{$\alpha_t$} & \textbf{$\sigma_t$} & \textbf{$t$-range} & \textbf{Training Objective} & \textbf{Sampling ODE} & \textbf{Initial Noise} \\
        \midrule
        EDM~\citep{edm,edm2} 
        & $1$ & $t$ & $[0,T]$ 
        & $\mathbb{E}\!\left[\|\hat{\bx}_\theta(\bx_t,t)-\bx_0\|_2^2\right]$ 
        & $\tfrac{d\bx_t}{dt}=\tfrac{\bx_t-\hat{\bx}_\theta(\bx_t,t)}{t}$ 
        & $\bx_T\!\sim\!\mathcal{N}(\mathbf{0},T^2\mathbf{I})$ \\
        \addlinespace[3pt]
        FM~\citep{flow-straight,flow-matching} 
        & $1-t$ & $t$ & $[0,1]$ 
        & $\mathbb{E}\!\left[\|\mathbf{v}_\theta(\bx_t,t) - (\bz-\bx_0)\|_2^2\right]$
        & $\tfrac{d\bx_t}{dt}=\mathbf{v}_\theta(\bx_t,t)$ 
        & $\bx_1\!\sim\!\mathcal{N}(\mathbf{0},\mathbf{I})$ \\
        \addlinespace[3pt]
        TrigFlow~\citep{albergobuilding,scm} 
        & $\cos t$ & $\sin t$ & $[0,\tfrac{\pi}{2}]$ 
        & $\mathbb{E}\!\left[\|\sigma_d F_\theta(\tfrac{\bx_t}{\sigma_d},t)-(\cos t\,\bz-\sin t\,\bx_0)\|_2^2\right]$ 
        & $\tfrac{d\bx_t}{dt}=\sigma_d F_\theta(\tfrac{\bx_t}{\sigma_d},t)$ 
        & $\bx_{\pi/2}\!\sim\!\mathcal{N}(\mathbf{0},\sigma_d^2\mathbf{I})$ \\
        \bottomrule
    \end{tabular}
\caption{Unified comparison of EDM, Flow Matching and TrigFlow formulations.}
\label{table_uni_form}
\end{table*}

\begin{itemize}
    \item \textbf{An Image-Free Distillation Framework.}
    We design a continuous-time consistency distillation framework that fully leverages the teacher model’s generative capability, enabling distillation to be performed entirely in the latent space without any image data. This image-free setting eliminates the need for VAE involvement and data preprocessing, making the process more efficient and lightweight.
    \item \textbf{A New Perspective on Distillation Samples.}  
    By systematically examining the forward-sampling strategy in sCM and the backward-sampling strategy in TBCM, we uncover the inherent differences between forward and backward sample spaces. This provides a novel understanding of how sampling strategies in different spaces influence the quality of consistency distillation.  
    \item \textbf{Low-Cost and High-Quality Distillation.}  
    The proposed framework significantly reduces GPU memory usage and shortens training time by approximately 40\%, while simultaneously improving latent consistency between training and inference. This ensures both low training cost and superior generation quality.  
\end{itemize}

%% file: sec/2_Related_Work.tex
\section{Related Work}
\label{sec:related}

\boldparagraph{Diffusion Models.} 
Diffusion models (DMs) have become a dominant paradigm in generative modeling since the introduction of DDPM~\citep{ddpm} and its improved variants~\citep{iddpm}. Numerous works, such as DDIM~\citep{ddim}, DPM-Solver~\citep{dpm-solver}, and EDM~\citep{edm}, have focused on accelerating and stabilizing the sampling process. Recently, Flow Matching~\citep{flow-straight, flow-matching} reformulated diffusion as learning continuous flows between data and noise, offering a unified perspective for deterministic generation. Building on these advances, SANA~\citep{sana} leverages the highly compressed DCAE~\citep{dcae} and a linear architecture to achieve efficient and high-quality generation.

\boldparagraph{Timestep Distillation.} 
Efforts to accelerate diffusion inference through timestep distillation fall into two main categories: trajectory-oriented and distribution-oriented approaches. 
Early trajectory-oriented methods~\citep{direct, progressive} leverage the teacher model’s full ODE trajectory to capture the mapping between noise and images. Consistency Models~\citep{cm, lcm, ctm, mcm, pcm, scm} further impose a self-consistency constraint, aligning $x_0$ predictions across adjacent timesteps.
Distribution-oriented approaches, in contrast, aim to match overall generative distributions. ADD~\citep{add} performs pixel-domain distillation with adversarial learning using pretrained perceptual encoders, whereas LADD~\citep{ladd} shifts this process into the latent space for computational efficiency. Variational Score Distillation (VSD)~\citep{dreamfusion, prolificdreamer} provides a non-adversarial alternative, with subsequent methods~\citep{dmd, dmd2, sid, sim} building on this idea to improve stability and effectiveness.

%% file: sec/3_Preliminaries.tex
\section{Preliminaries}
\label{sec:preliminaries}

\subsection{Different Formulations of Diffusion Models}
\label{sec:formulation}
Diffusion-based generative models synthesize data by reversing a progressive noising process. Given $\bx_0 \sim p_{\mathrm{data}}$, a perturbed sample is defined as $\bx_t = \alpha_t \bx_0 + \sigma_t \bz$, where $\bz \sim \mathcal{N}(\mathbf{0}, \mathbf{I})$ and $(\alpha_t, \sigma_t)$ defines the noise schedule.
Different parameterizations yield distinct formulations (see Tab.~\ref{table_uni_form}), mainly differing in interpolation schedules and vector field parameterization.

\subsection{Continuous-Time Consistency Models}
\label{sec:ctcm}
Consistency Models (CMs)~\citep{cm} learn to predict the clean data $\bx_0$ from an arbitrary noisy observation $\bx_t$ along the trajectory of a probability flow ODE.
Formally, a CM parameterizes a neural network $\boldsymbol{f_{\theta}}(\bx_t, t)$ that outputs the estimated clean signal, which remains consistent across different noise levels.

\boldparagraph{From Discrete to Continuous.}
Early CMs~\citep{cm, lcm} employ discrete-time training with a consistency loss between neighboring timesteps:
\begin{equation}
    l_{CM}^{\Delta t} = \mathbb{E}_{\bx_t, t}\!\left[d(\boldsymbol{f_{\theta}}(\bx_t, t), \boldsymbol{f_{\theta^-}}(\bx_{t-\Delta t}, t-\Delta t))\right],
\label{eq:disc cm loss}
\end{equation}
where $d(\cdot,\cdot)$ is a distance metric. This discrete formulation inevitably introduces discretization errors.
Continuous-Time Consistency Models~\citep{cm,scm} overcome this by taking the infinitesimal limit $\Delta t \rightarrow 0$, yielding a smooth training objective free of discretization artifacts:
\begin{equation}
    \small l_{CM}^{cont.} = \mathbb{E}_{\bx_t, t}\!\left[w(t)\,\big\langle 
    \boldsymbol{f_{\theta}}(\bx_t, t), 
    \tfrac{\mathrm{d} \boldsymbol{f_{\theta^-}}}{\mathrm{d} t}(\bx_t, t)
    \big\rangle\right].
\label{eq:continuous cm loss}
\end{equation}

\begin{figure*}[th]
    \centering
    \includegraphics[width=\textwidth]{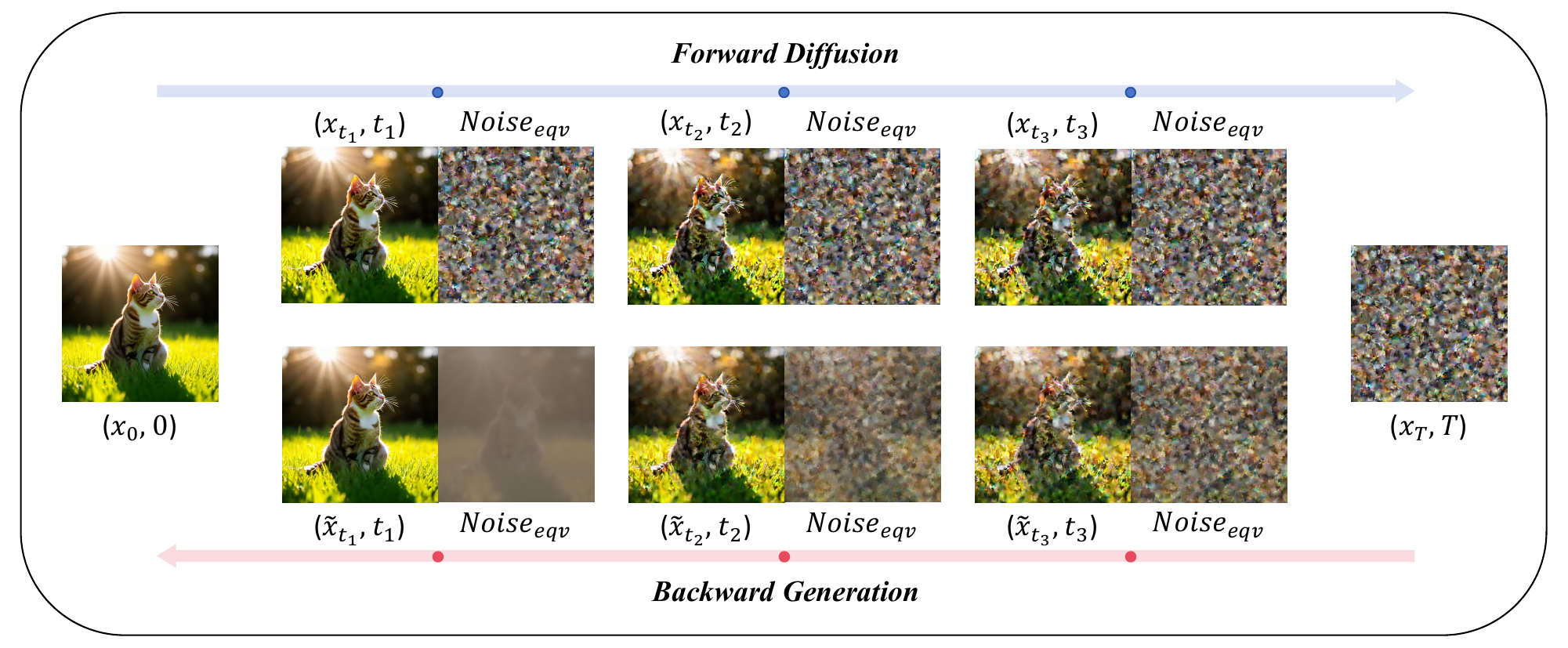}
    \caption{
    \textbf{Discrepancy of Equivalent Noise Between Forward and Backward Processes.} The equivalent noise (see ~\eqref{eq:eqv noise}) remains constant in forward diffusion, but evolves noticeably in backward generation, reflecting the training–inference inconsistency.
    }
    \label{fig:for-back}
\end{figure*}

\boldparagraph{Trigonometric Parameterization.}
In the classical Flow Matching framework, the term $\frac{\mathrm{d} \boldsymbol{f_{\theta^-}}(\bx_t, t)}{\mathrm{d} t}$ in ~\eqref{eq:continuous cm loss} can be expressed as
\begin{equation}
    \frac{\mathrm{d} \boldsymbol{f_{\theta^-}}(\bx_t, t)}{\mathrm{d} t}
    = \frac{\partial \boldsymbol{f_{\theta^-}}(\bx_t, t)}{\partial t}
    + \nabla_{\bx_t} \boldsymbol{f_{\theta^-}}(\bx_t, t) \frac{\mathrm{d}\bx_t}{\mathrm{d}t}.
\label{eq:fm dfdt}
\end{equation}
However, previous work~\citep{cm, ect} found that this optimization objective is highly unstable and difficult to scale up for large models or datasets. To address this issue, the TrigFlow architecture was proposed (see Sec.~\ref{sec:formulation}). Under the TrigFlow formulation, this term is instead represented as
\begin{equation}
    \begin{aligned}
        \tfrac{\mathrm{d} \boldsymbol{f_{\theta^-}}(\bx_t, t)}{\mathrm{d} t}
        =& -\cos(t)\Big(\sigma_d \boldsymbol{F_{\theta^-}}\!\left(\tfrac{\bx_t}{\sigma_d}, t\right)
         - \tfrac{\mathrm{d}\bx_t}{\mathrm{d}t}\Big)\\
        & -\sin(t)\Big(\bx_t + \sigma_d \tfrac{\mathrm{d} \boldsymbol{F_{\theta^-}}\!\left(\tfrac{\bx_t}{\sigma_d}, t\right)}{\mathrm{d} t}\Big).
    \end{aligned}
\label{eq:trig dfdt}
\end{equation}

\boldparagraph{Aligning FM and Trig Parameterizations.}
Sana-Sprint~\citep{sana-sprint} unifies the FM and Trig representations through explicit transformations.
The time and data mapping is
\begin{equation}
    t_{\texttt{FM}} = \frac{\sin(t_{\texttt{Trig}})}{\sin(t_{\texttt{Trig}}) + \cos(t_{\texttt{Trig}})},
\label{eq:t trans}
\end{equation}
\begin{equation}
    \bx_{t,\texttt{FM}} = \frac{\bx_{t,\texttt{Trig}}}{\sigma_d}\cdot\sqrt{t_{\texttt{FM}}^2 + (1-t_{\texttt{FM}})^2}.
\label{eq:xt trans}
\end{equation}
The output transformation is
\begin{equation}
    \small
    \begin{aligned}
        &\widehat{\boldsymbol{F_{\theta}}}\left(\frac{\bx_{t,\texttt{Trig}}}{\sigma_d}, t_{\texttt{Trig}}, \boldsymbol{y}\right) \\[3pt]
        = &\frac{1}{\sqrt{t_{\texttt{FM}}^2 + (1-t_{\texttt{FM}})^2}} \Big[
          (1-2t_{\texttt{FM}}) \bx_{t, \texttt{FM}} \\
          &\quad + (1-2t_{\texttt{FM}} + 2t_{\texttt{FM}}^2) \boldsymbol{v_{\theta}}(\bx_{t, \texttt{FM}}, t_{\texttt{FM}}, \boldsymbol{y})
        \Big].
    \end{aligned}
\label{eq:output trans}
\end{equation}

%% file: sec/4_Methods.tex
\section{Method}
\label{sec:methods}

\subsection{Trajectory-Backward Consistency Models}
\label{sec:tbcm}

\boldparagraph{\textit{Finding 1: Resource Bottlenecks in Distillation.}}
During the distillation process, VAE encoding constitutes a major source of GPU memory consumption, while prompt encoding occupies a substantial portion of the training time. 

\begin{figure}[h]
    \centering
    \includegraphics[width=0.5\textwidth]{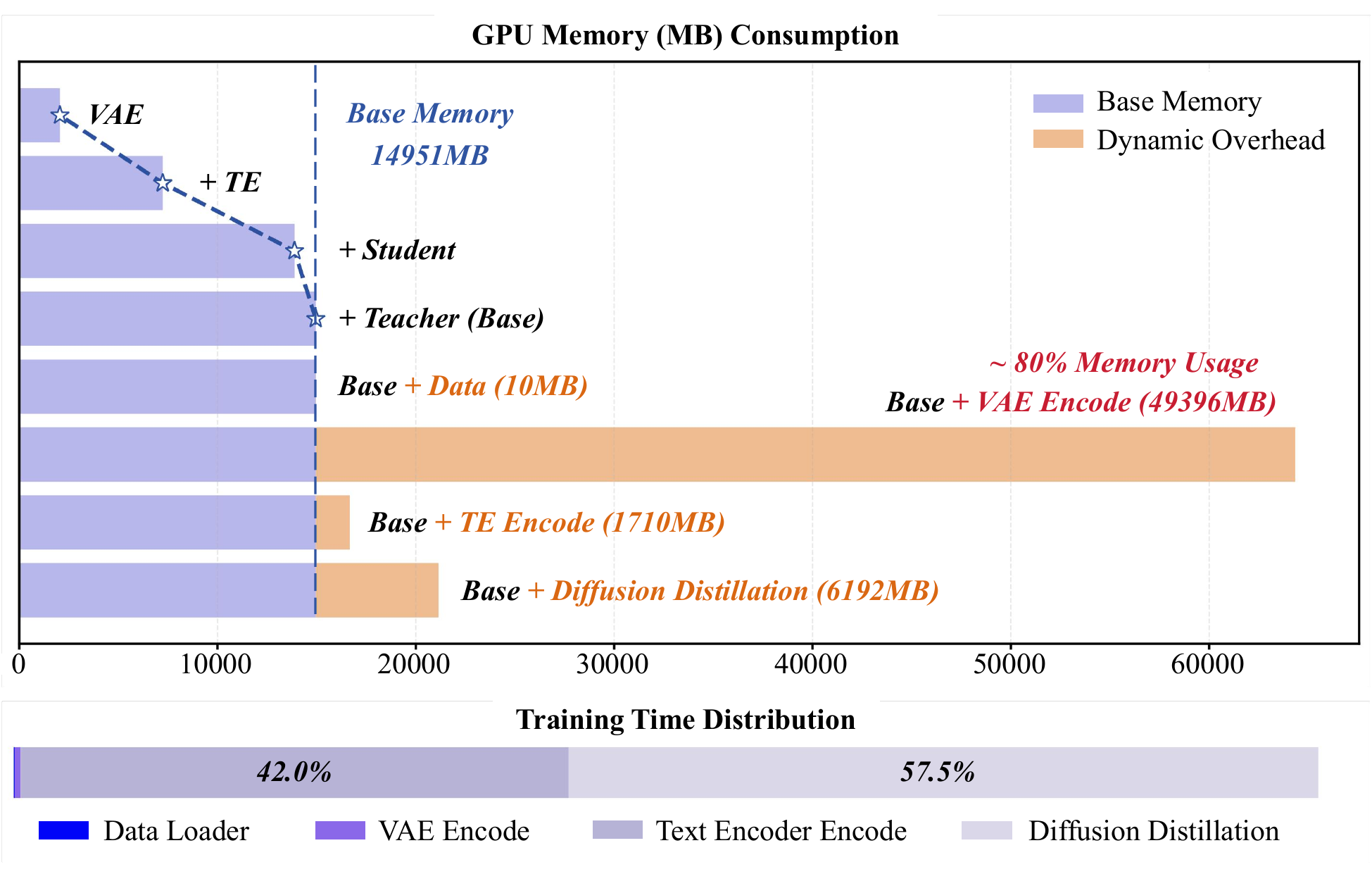}
    \caption{\textbf{Resource Bottlenecks in Continuous-Time Consistency Distillation.} \textit{\textbf{Top:}} Memory usage breakdown during distillation. \textit{\textbf{Bottom:}} Training time breakdown during distillation.}
    \label{fig:bottlenecks}
\end{figure}

As shown in Fig.~\ref{fig:bottlenecks}, the top part illustrates the memory usage breakdown during distillation, where \textit{Base Memory} consists of the VAE, Text Encoder (TE), Student and Teacher models, and \textit{Dynamic Overhead} denotes the maximum memory consumption across different stages. The VAE encoding stage exhibits significantly higher usage than others, accounting for approximately 80\% of the total memory consumption. 
The bottom part shows the time breakdown during distillation, where the Data Loader and VAE Encoding stages account for less than 1\% of the total time, while the Text Encoder contributes a substantial proportion, comparable to the Diffusion Distillation process.

\begin{figure*}[th]
    \centering
    \includegraphics[width=\textwidth]{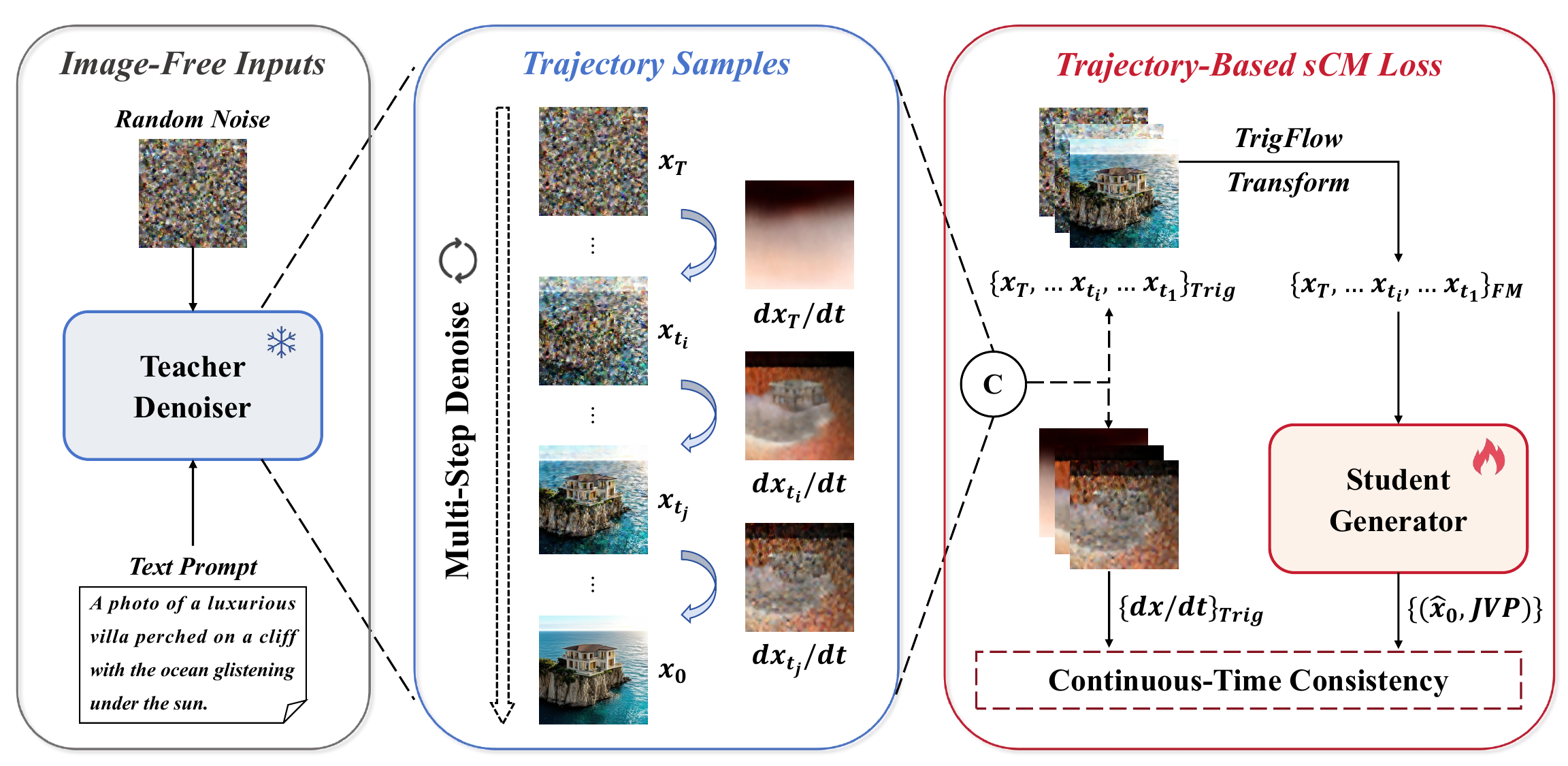}
    \caption{\textbf{Distillation Paradigm of TBCM.} 
    \textit{\textbf{Left:}} Distillation begins with random noise and text prompt inputs. \textit{\textbf{Middle:}} Multiple samples are generated for a single prompt within the latent space. \textit{\textbf{Right:}} The collected samples are used to compute the consistency loss.}
    \label{fig:scheme}
\end{figure*}

\textbf{\textit{Insight:}} 
To overcome the GPU memory bottleneck during distillation, we can fully leverage the generative capability of the pretrained model and perform distillation purely in the latent space. 
This design decouples the distillation process from the VAE encoder, establishing an \textit{image-free} distillation paradigm that fundamentally eliminates dependence on the VAE.
To mitigate the training-time bottleneck, generating multiple samples for a single prompt can effectively amortize the text encoding overhead, thereby accelerating the overall training process.

\boldparagraph{\textit{Finding 2: Training--Inference Inconsistency.}}
In diffusion model distillation, the samples received during training differ substantially from those encountered during inference. 

Although diffusion models are pretrained under a \textit{forward sampling} paradigm, where training samples are obtained by adding noise of varying magnitudes to clean images, they perform inference along a fundamentally different \textit{backward sampling} trajectory. 
To characterize this discrepancy, we introduce the concept of Equivalent Noise:
\begin{equation}
    Noise_{eqv} = \frac{\bx_t - \cos(t) \hat{\bx}_0}{\sin(t)}.
\label{eq:eqv noise}
\end{equation}

As shown in Fig.~\ref{fig:for-back}, we observe that the equivalent noise remains consistent during the forward process, while exhibiting significant shifts during the backward process. Specifically, it gradually transforms from random noise to patterns correlated with the prediction target, revealing that diffusion models learn a coarse-to-fine paradigm, which aligns with observations from recent studies~\citep{oms-dpm, prompt-to-prompt, delta-dit}.

To demonstrate that this discrepancy is not merely instance-level but manifests systematically across the distribution, we further visualize the overall sample distribution using t-SNE (see Appendix). 
The results consistently show substantial inconsistency between the sample distributions in the forward and backward processes.

Such a training--inference inconsistency indicates that the constraints applied to noisy samples during distillation are not properly aligned with the actual inference trajectory, which may potentially undermine the effectiveness of the distillation process.

\textbf{\textit{Insight:}} To mitigate the discrepancy between training and inference, distillation should align training samples with the actual backward trajectory by sampling along the pretrained model’s inference path, thereby enhancing the effectiveness of the distillation process.

\boldparagraph{\textit{Solution: Trajectory-Driven Consistency Learning.}}
To jointly mitigate the \textit{resource bottlenecks} and the \textit{training–inference inconsistency} observed in previous sCM distillation frameworks, we introduce a trajectory-based distillation scheme that operates directly in the latent space without invoking the VAE encoder, while simultaneously generating multiple samples along the trajectory for each prompt.

Specifically, instead of generating noisy inputs by adding noise to VAE-encoded images, we explicitly simulate the teacher’s denoising trajectory as
\begin{equation}
    \begin{aligned}
    \frac{d\bx_t}{dt} &= F_{teacher}\left(\frac{\bx_t}{\sigma_d}, t\right), \\
    \bx_{t-\Delta t} &= \cos(\Delta t)\,\bx_t - \sin(\Delta t)\sigma_d\frac{d\bx_t}{dt}.
    \end{aligned}
\label{eq:trajectory sample}
\end{equation}
By integrating this ODE trajectory, we obtain both the intermediate states $\bx_t$ and the teacher-predicted temporal derivatives $\frac{d\bx_t}{dt}$, which are essential for the subsequent distillation while avoiding repeated VAE encoding. These quantities are then used to compute $\widehat{\boldsymbol{F_{\theta}}}$ in ~\eqref{eq:trig dfdt}, and further serve to construct the continuous-time consistency loss defined in ~\eqref{eq:continuous cm loss}. The overall framework is illustrated in Fig.~\ref{fig:scheme}.

\subsection{Sampling Schemes Shaping the Sample Space}
\label{sec:t_sample}

\boldparagraph{\textit{Finding 3: Sample Space Drives Consistency Distillation.}}
From the composition of the sCM loss (see ~\eqref{eq:trig dfdt}), it is evident that the samples affecting sCM training depend not directly on the clean image \(x_0\), but rather on the noisy samples \(x_t\). Pairs of \((x_t, t)\) constitute the complete sample space for sCM training.

Following the discussion in Section.~\ref{sec:tbcm} on training--inference inconsistency, we define the sample space obtained from forward sampling as the diffusion space, and that from backward sampling as the generation space. 

\textbf{\textit{Insight:}} The composition of the sample space is a decisive factor affecting the effectiveness of consistency distillation. By flexibly adjusting the sample scheme, we can achieve optimal distillation quality.

\boldparagraph{\textit{Preliminary: Sampling Scheme in Diffusion Space.}}
Conventional sCM and Sana-Sprint methods perform sampling in the diffusion space. 
In diffusion space, since the clean image \(\bx_0\) is fixed, the training samples \((\bx_t, t)\) are influenced only by the sampling timestep \(t\) and the random noise. 
Consequently, the corresponding sampling strategy typically focuses on the distribution of noise magnitudes, i.e., the distribution of sampled timesteps.

A widely used approach is the logit-Normal proposal distribution. 
In the TrigFlow architecture, it is used to sample \(\tan(t)\), such that $e^{\sigma_d \tan(t)} \sim \mathcal{N}(P_\text{mean}, P_\text{std}^2)$, which is adopted in both sCM and Sana-Sprint.

\boldparagraph{\textit{Extension: Sampling Scheme in Generation Space.}}
Our proposed TBCM performs sampling in the generation space. 
In generation space, the training samples \((\bx_t, t)\) are influenced not only by the sampling timestep \(t\) but also by the inference trajectory—that is, by each intermediate timestep along the path from pure noise to the target sample.
For a given number of sampling steps \(N\), the \(i\)-th training sample along the trajectory is affected by \(\{t_{N-1}, \dots, t_{i+1}, t_i\}\). 
Therefore, the choice of trajectory can significantly impact the training outcome. 

Consequently, when designing the sampling strategy in the generation space, we should consider not only the overall distribution of timesteps \(t\) but also the manner in which the sampling trajectory is obtained. 
Relevant ablation studies are presented in Sec.~\ref{sec:ablation}.

\subsection{Additional Adjustments}
\label{sec:additional}
\boldparagraph{Brightness Filter.}
Due to the uncertainty in the sampling trajectory, the teacher model may generate low-quality predictions of the clean image \(\hat{\bx_0}\) after completing the entire trajectory. 
We observe that these low-quality images often share a common characteristic: they exhibit low overall brightness. 
To filter such samples without involving a VAE, We observe that dark images in pixel space are mapped to latent representations that are close to those of an all-black image.
This property allows us to directly filter low-brightness samples in the latent space (see Appendix).

\boldparagraph{Stability Hyperparameter.}
In sCM and Sana-Sprint, to stabilize the unstable term
\(\sin(t)(\bx_t + \sigma_d \frac{d F_{\theta^-}}{d t})\) in \(\frac{d F_{\theta^-}}{d t}\), 
the factor \(\sin(t)\) is replaced with \(r \cdot \sin(t)\), and \(r\) gradually warms up from 0 to 1 during early training. 
In TBCM, we find that 1.0 is not the optimal value for \(r\). 
We therefore explore different choices of \(r\) as well as various schedules for its change. 
Experimental details are presented in Sec.~\ref{sec:ablation}.

\begin{figure}[h]
    \centering
    \includegraphics[width=0.5\textwidth]{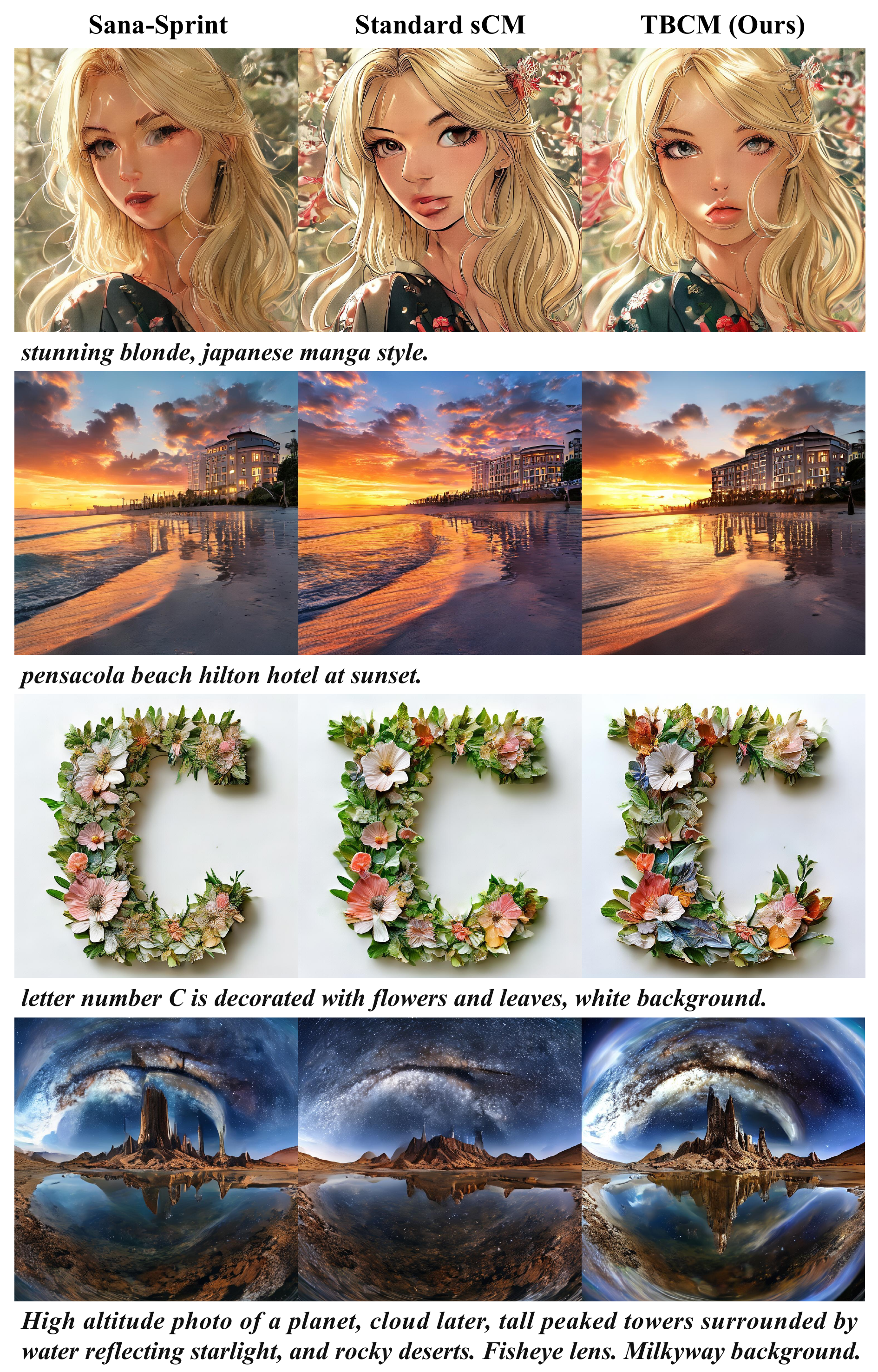}
    \caption{Visual Comparison under One-Step Generation.}
    \label{fig:compare}
\end{figure}

%% file: sec/5_Experiments.tex
\section{Experiments}
\label{sec:experiments}

\begin{table*}[th]
	\centering
	\begin{tabular}{cccccccc}
		\toprule
        \multicolumn{2}{c}{\multirow{2}{*}{\textbf{Methods}}} & \textbf{Inference} & \textbf{Throughput} & \textbf{Latency} & \textbf{Params} & \multirow{2}{*}{\textbf{FID~$\downarrow$}} & \multirow{2}{*}{\textbf{CLIP~$\uparrow$}} \\
        \multicolumn{2}{c}{} & \textbf{steps} & \textbf{(samples/s)} & \textbf{(s)} & \textbf{(B)} &  &\\
		\midrule
        \multicolumn{1}{c}{\multirow{3}{*}{{\textbf{Pre-train}}}}
        & SDXL~\citep{sdxl}                       & 50 & 0.15 & 6.5  & 2.6 & 6.63             & 29.03              \\
        & PixArt-$\Sigma$~\citep{pixart-sigma}    & 20 & 0.4  & 2.7  & 0.6 & 6.15             & 28.26              \\
        & SANA~\citep{sana}                       & 20 & 1.7  & 0.9  & 0.6 & 5.81             & 28.36              \\
        \midrule
        \multicolumn{1}{c}{\multirow{8}{*}{{\textbf{Distillation}}}}
        & SDXL-LCM~\cite{lcm}                       & 1  & 3.36 & 0.32 & 2.6 & 50.51            & 24.45              \\
        & PixArt-LCM~\cite{pixart-delta}          & 1  & 4.26 & 0.25 & 0.6 & 73.35            & 23.99              \\
        & PixArt-DMD~\cite{pixart-sigma}          & 1  & 4.26 & 0.25 & 0.6 & 9.59             & 26.98              \\
        & PCM~\cite{pcm}                          & 1  & 3.16 & 0.40 & 2.8 & 30.11            & 26.47              \\
        & SDXL-DMD2~\cite{dmd2}                     & 1  & 3.36 & 0.32 & 2.6 & 7.10             & \textbf{28.93}     \\
        & Sana-Sprint~\cite{sana-sprint}          & 1  & 7.22 & 0.21 & 0.6 & \underline{7.04} & 28.04              \\
        & sCM~\cite{scm}                          & 1  & 7.22 & 0.21 & 0.6 & 7.46             & 27.74              \\
        & \textbf{TBCM (Ours) }                   & 1  & 7.22 & 0.21 & 0.6 & \textbf{6.52}    & \underline{28.08}  \\
		\bottomrule
        
	\end{tabular}
\caption{\textbf{Comparison of our method with various approaches on the MJHQ-30k test set.} The reported results of baseline methods are mainly sourced from the Sana-Sprint report~\citep{sana-sprint}.}

\label{table_model_comparasion}
\end{table*}

\begin{table*}[ht]
	\centering
	\begin{tabular}{cccccc}
		\toprule
        \multicolumn{2}{c}{\multirow{2}{*}{\textbf{Methods}}} & \textbf{Average Sample Time} & \textbf{Total Training Samples} & \textbf{Memory Usage} & \textbf{Total Training Time} \\ 
        \multicolumn{2}{c}{} & \textbf{(GPU * s)} & \textbf{(BS * Steps)} & \textbf{(GB)} & \textbf{(GPU * h)} \\
        \midrule
        & Sana-Sprint~\cite{sana-sprint}            & 2.7          & 512 * 20000 & 72.0 \textcolor{BrickRed}{(+14.5\%)}         & 7680 \textcolor{ForestGreen}{(-3.60\%)}         \\
        & sCM~\cite{scm}                    & 2.8          & 512 * 20000 & 62.9 \textcolor{gray}{(\phantom{-}0.00\%)}         & 7965 \textcolor{gray}{(\phantom{-}0.00\%)}         \\
        & \textbf{TBCM (Ours)}   & \textbf{1.6} & 512 * 20000 & \textbf{22.6} \textcolor{ForestGreen}{(-64.1\%)} & \textbf{4640} \textcolor{ForestGreen}{(-41.7\%)} \\
		\bottomrule
	\end{tabular}
\caption{\textbf{Training costs comparison of different schemes.} All training time measurements were conducted on a cluster with 4 nodes (32 NVIDIA V100 GPUs in total), while memory usage was evaluated on a single A100 GPU with a batch size of 16.}

\label{table_training_cost}
\end{table*}

\subsection{Main Results}
\label{sec:main_results}
\boldparagraph{Experimental Setup.}
Thanks to the \textit{image-free} nature of our distillation pipeline, we directly collect 1M randomly sampled text prompts for training without any paired image data. All experiments are conducted on a cluster of 32 NVIDIA V100 GPUs (32 GB each). 
For a fair comparison, we strictly follow the training configurations of Sana-Sprint except for a minor learning rate adjustment introduced by the trajectory sampling scheme. The teacher model used in our distillation is the officially released Sana-Sprint 0.6B teacher model. 
We evaluate all models on the MJHQ-30k~\citep{playground} benchmark using FID~$\downarrow$ (Fréchet Inception Distance) and CLIP Score~$\uparrow$ metrics to measure perceptual quality and text--image alignment.

\boldparagraph{Results and Analysis.}
As shown in Tab.~\ref{table_model_comparasion}, our proposed TBCM achieves a remarkable balance between efficiency and fidelity in the one-step generation setting. Specifically, TBCM obtains an outstanding \textbf{FID of 6.52} and a \textbf{CLIP score of 28.08}, outperforming existing distillation-based methods, including Sana-Sprint (7.04 FID, 28.04 CLIP score), under the same training setup. 
As shown in Tab.~\ref{table_training_cost}, our TBCM method reduces training costs by over 40\% and saves more than 60\% of GPU memory compared to Sana-Sprint and sCM.
Fig.~\ref{fig:compare} shows a comparison of the visualization results of Sana-Sprint, sCM, and TBCM.
The results validate that our method successfully leverages trajectory-sampled pairs to transfer teacher knowledge more effectively, leading to sharper visual quality and stronger text--image consistency in a single inference step.

\subsection{Ablation Study}
\label{sec:ablation}

\boldparagraph{Sampling Schemes.}
As discussed in Sec.~\ref{sec:t_sample}, the sampling strategy strongly affects the distribution of trajectory samples in the generation space. 
To verify this, we perform ablation experiments comparing three sampling schemes:

\begin{figure}[h]
    \centering
    \includegraphics[width=0.5\textwidth]{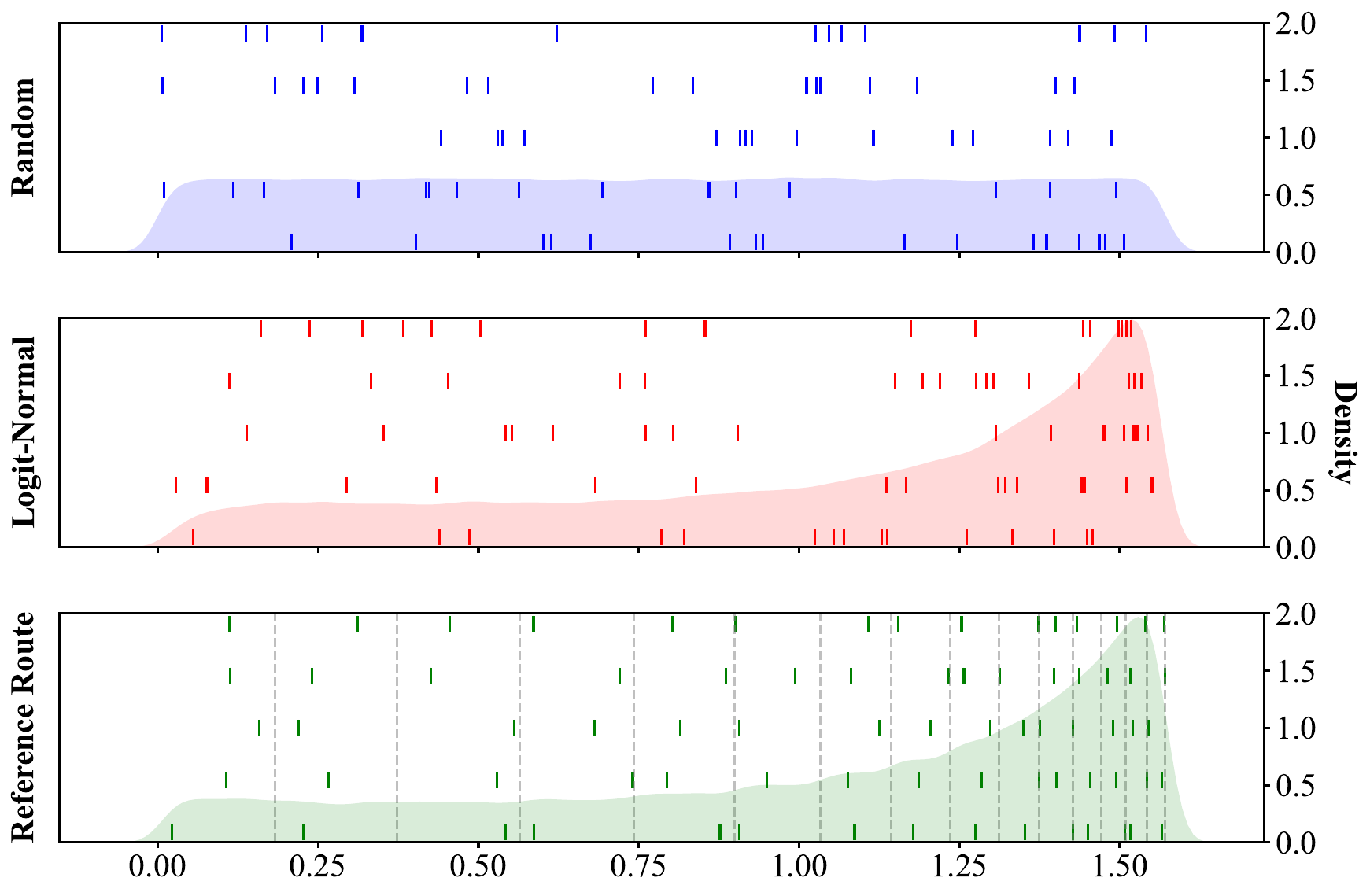}
    \caption{\textbf{Sampling Patterns of Different Strategies.} 
    \textbf{\textit{Top to bottom:}} Three sampling strategies — Random, Logit-Normal, Reference Route. 
    \textbf{\textit{Vertical lines:}} Five random sampling instances per strategy. 
    \textbf{\textit{Shaded regions:}} Sampling density distributions.}
    \label{fig:sample_scheme}
\end{figure}

\begin{itemize}
    \item \textbf{Random:} uniformly sample $N$ points from the interval $t \in [0, \pi/2]$ to form the denoising trajectory.
    \item \textbf{Logit-Normal:} adopt the common logit-normal sampling in diffusion space, following the same hyperparameters as Sana-Sprint ($P_{\text{Mean}} = 0.2$, $P_{\text{Std}} = 1.6$).
    \item \textbf{Reference Route:} extract the timesteps from a Flow-Euler Scheduler incorporating a schedule shift, map them to the $[0, \pi/2]$ interval via the inverse of ~\eqref{eq:t trans}, and use the resulting trajectory as a reference for partitioned sampling, which allocates timesteps to different partitions to balance sampling density and preserve trajectory fidelity.
\end{itemize}

As shown in Tab.~\ref{table_ablation_schemes}, the Logit-Normal strategy improves upon the Random scheme by optimizing the timestep distribution, while the Reference Route approach further constrains each trajectory to include samples within every subregion, thereby reducing randomness compared to probabilistic sampling. 
Although the timestep distributions of the Logit-Normal and Reference Route strategies are visually similar (Fig.~\ref{fig:sample_scheme}), the Reference Route strategy achieves the best FID and CLIP performance, followed by the Logit-Normal one. 
These results support our claim in Sec.~\ref{sec:t_sample} regarding the influence of generation-space sampling.

\begin{table}[h]
	\centering
	\begin{tabular}{ccc}
		\toprule
        \textbf{Sample Schemes} & \textbf{FID~$\downarrow$} & \textbf{CLIP~$\uparrow$} \\
        \midrule
        Random          & 9.23          & 27.74          \\
        Logit-Normal      & 7.18          & 27.80          \\
        Reference Route & \textbf{6.79} & \textbf{28.00} \\
		\bottomrule
	\end{tabular}
\caption{Ablation study on different trajectory sampling methods.}
\label{table_ablation_schemes}
\end{table}

\boldparagraph{Sampling Steps.}
Beyond the sampling scheme, the number of sampled steps also significantly impacts the distillation outcome, as it determines the coverage of the generation trajectory. 
As shown in Tab.~\ref{table_ablation_steps}, increasing the number of sampled steps leads to a clear improvement in FID, while the CLIP score remains relatively stable. 
This observation aligns with our finding that fewer steps yield lower-quality clean samples in the denoising trajectory.

\begin{table}[ht]
	\centering
	\begin{tabular}{ccccc}
		\toprule
        \textbf{Sample Steps} & \textbf{Memory Usage} & \textbf{FID~$\downarrow$} & \textbf{CLIP~$\uparrow$} \\
        \midrule
        8  & 20.4 GB & 7.52             & 27.98          \\
        12 & 21.2 GB & 6.87             & \textbf{28.00} \\
        16 & 21.9 GB & 6.77             & \textbf{28.00} \\
        20 & 23.6 GB & \textbf{6.58}    & 27.98          \\
		\bottomrule
        
	\end{tabular}
\caption{Ablation study on the number of sampling steps.}
\label{table_ablation_steps}
\end{table}

\boldparagraph{Hyperparameter $\boldsymbol{R}$.}
We further analyze the stability-related hyperparameter $R$ introduced in Sec.~\ref{sec:additional}. 
As shown in Tab.~\ref{table_ablation_r}, setting the final value $R_{\text{final}}=0.75$ achieves the most stable and effective training. 
Tab.~\ref{table_ablation_r_var} further compares two scheduling strategies, revealing that the \textit{Warmup–Cooldown} schedule (i.e., warming up to~1 and then cooling down to~$R_{\text{final}}$) outperforms the direct \textit{Warmup-to-$R_{\text{final}}$} scheme. 

\begin{table}[ht]
	\centering
	\begin{tabular}{ccc}
		\toprule
        \textbf{Final $\boldsymbol{R}$ Value} & \textbf{FID~$\downarrow$} & \textbf{CLIP~$\uparrow$} \\
        \midrule
        1.0  & 7.66          & 28.00          \\
        0.75 & \textbf{6.77} & \textbf{28.00} \\
        0.65 & 6.81          & 27.98          \\
        0.5  & 7.41          & 27.98          \\
		\bottomrule
	\end{tabular}
\caption{Ablation study on the selection of $R_{\text{final}}$ value.}
\label{table_ablation_r}
\end{table}

\begin{table}[ht]
	\centering
	\begin{tabular}{ccc}
		\toprule
        \textbf{$\boldsymbol{R}$ Schedule} & \textbf{FID~$\downarrow$} & \textbf{CLIP~$\uparrow$} \\
        \midrule
        Warmup   &   6.58            &  27.98             \\
        Warmup–Cooldown                &   \textbf{6.48}   &  \textbf{28.00}    \\
		\bottomrule
	\end{tabular}
\caption{Ablation study on different $R$ scheduling strategies.}
\label{table_ablation_r_var}
\end{table}

%% file: sec/6_Conclusion.tex
\section{Conclusion}
\label{sec:conclusion}

In this work, we presented TBCM, a continuous-time consistency distillation framework that conducts the entire distillation procedure within the latent space under image-free conditions. By removing VAE involvement, TBCM substantially reduces training cost and enables efficient, scalable deployment across a wide range of diffusion backbones. Furthermore, by bridging the training–inference inconsistency, TBCM achieves strong one-step generation performance despite its lightweight design.

Despite these advantages, TBCM still exhibits certain limitations. Without real image supervision, its effectiveness is inherently constrained by the capacity and biases of the teacher model. Imperfect generative behaviors from the teacher may propagate to the student, potentially limiting sample diversity or inducing mild mode collapse. This highlights a key challenge in image-free distillation: the student’s performance is tightly coupled to the quality of the teacher’s synthetic trajectories.

From a broader perspective, our formulation introduces the notion of sample space as a supplement to consistency distillation. The expressiveness and structure of the constructed sample space play a critical role in shaping the behavior of continuous-time distillation. Designing more expressive, well-structured sample spaces therefore remains an open and valuable research direction.

We believe that future work combining TBCM with complementary generative or regularization strategies may further mitigate teacher-induced limitations. More broadly, we hope the conceptual lens of sample space inspires further research into more fundamental, principled, and generalizable formulations of consistency distillation.

%% file: sec/X_suppl.tex
\clearpage
\setcounter{page}{1}

\setcounter{section}{0}
\renewcommand{\thesection}{\Alph{section}}

\maketitlesupplementary

This supplementary material provides additional analyses and results to complement the main paper. Specifically, Sec.~\ref{sup:equ noise} visualizes the distribution shift of equivalent noise during the diffusion and generation processes. Sec.~\ref{sup:sampling} compares the effects of different sampling strategies on the predicted $\bx_0$. Sec.~\ref{sup:algorithm} presents the full TBCM algorithm. Sec.~\ref{sup:brightness} illustrates the Brightness Filter for identifying low-quality latent samples, and Sec.~\ref{sup:generation} shows additional multi-step generation results.

\newcommand{\algohighlight}[1]{\textcolor{blue}{#1}}
\algrenewcommand\algorithmicfor{\textcolor{blue}{\textbf{for}}}
\algrenewcommand\algorithmicdo{\textcolor{blue}{\textbf{do}}}
\algrenewcommand\algorithmicend{\textcolor{blue}{\textbf{end}}}
\begin{algorithm*}[t]
\caption{Training Algorithm of TBCM.}\label{alg:TBCM}
\centering
\begin{algorithmic}
\State \textbf{Input:} \algohighlight{prompt dataset $\gD$}, pretrained diffusion model $\F_{\text{pretrain}}$ with parameter $\theta_{\text{pretrain}}$, model $\F_\theta$, weighting $w_\phi$, learning rate $\eta$, constant $c$, warmup iteration $H$, \algohighlight{black latents $\z_b$, final r value $r_f$, calmdown start iteration $S_r$, calmdown steps $T_r$}.
\State \textbf{Note:} \algohighlight{$\sigma_d$ is not required in TBCM but is kept here for notational consistency.}
\State \textbf{Init:} $\theta \gets \theta_{\text{pretrain}}$, $\text{Iters}\gets 0$.
\Repeat
    \State \algohighlight{$\y\sim \gD$, $\x_t\sim \gN(\vzero,\sigma_d^2\mI)$ \Comment{Image-free inputs}}
    \State \algohighlight{$\mathcal{X} \gets \emptyset$, $\mathcal{V} \gets \emptyset$}
    \State \algohighlight{get denoise trajectory $\mathcal{T}$ from sampling schemes}
    \For{\algohighlight{each timestep $t_i$ in trajectory $\mathcal{T}$}}
        \State \algohighlight{$\frac{\dm \x_t}{\dm t} \gets \sigma_d\F_{\text{pretrain}}(\frac{\x_t}{\sigma_d},t,\y)$}
        \State \algohighlight{$\mathcal{X} \gets \mathcal{X} \cup \{\x_t\}$, $\mathcal{V} \gets \mathcal{V} \cup \{\frac{\dm \x_t}{\dm t}\}$}
        \State \algohighlight{$\x_t \gets \cos(t_i - t_{i+1})\x_t - \sin(t_i - t_{i+1})\frac{\mathrm{d}\x_t}{\mathrm{d}t}$}
    \EndFor
    \State \algohighlight{$mask \gets \text{black\_filter}(\x_t / \sigma_\text{data},\z_b)$  \Comment{Brightness filter}}
    \State \algohighlight{$\x_t \gets \text{Concatenate}(\mathcal{X})$, $\frac{\dm \x_t}{\dm t} \gets \text{Concatenate}(\mathcal{V})$, $t \gets \mathcal{T}$  \Comment{Trajectory sampling}}
    
    \State $r \gets \min(1, \text{Iters} / H)$ \Comment{Tangent warmup}
    \State \algohighlight{$p \gets \min(\max((\text{Iters} - S_r) / T_r, 0), 1)$}
    \State \algohighlight{$r \gets (1-p)\cdot r + p\cdot r_f$ \Comment{$\mathcal{R}$ adjustment}}
    \State $\vg \gets -\cos^2(t)(\sigma_d\F_{\theta^-} - \frac{\dm\x_t}{\dm t}) - r\cdot \cos(t)\sin(t) (\x_t + \sigma_d \frac{\dm \F_{\theta^-}}{\dm t})$  \Comment{JVP rearrangement}
    \State $\vg \gets \vg / (\|\vg\| + c)$ \Comment{Tangent normalization}
    \State $\gL(\theta,\phi) \gets \frac{e^{w_\phi(t)}}{D}\|
        \F_\theta(\frac{\x_t}{\sigma_d},t,y) - \F_{\theta^-}(\frac{\x_t}{\sigma_d},t,y)
        - \vg
    \|_2^2 - w_\phi(t)$ \Comment{Adaptive weighting}
    \State \algohighlight{$\gL(\theta,\phi) \gets mask\cdot \gL(\theta,\phi)$}
    \State $(\theta, \phi) \gets (\theta, \phi) - \eta \nabla_{\theta,\phi}\gL(\theta,\phi)$
    \State $\text{Iters}\gets \text{Iters} + 1$
\Until convergence
\end{algorithmic}
\end{algorithm*}

\section{Distribution Shift of Equivalent Noise.}
\label{sup:equ noise}
\begin{figure}[h]
    \centering
    \includegraphics[width=0.5\textwidth]{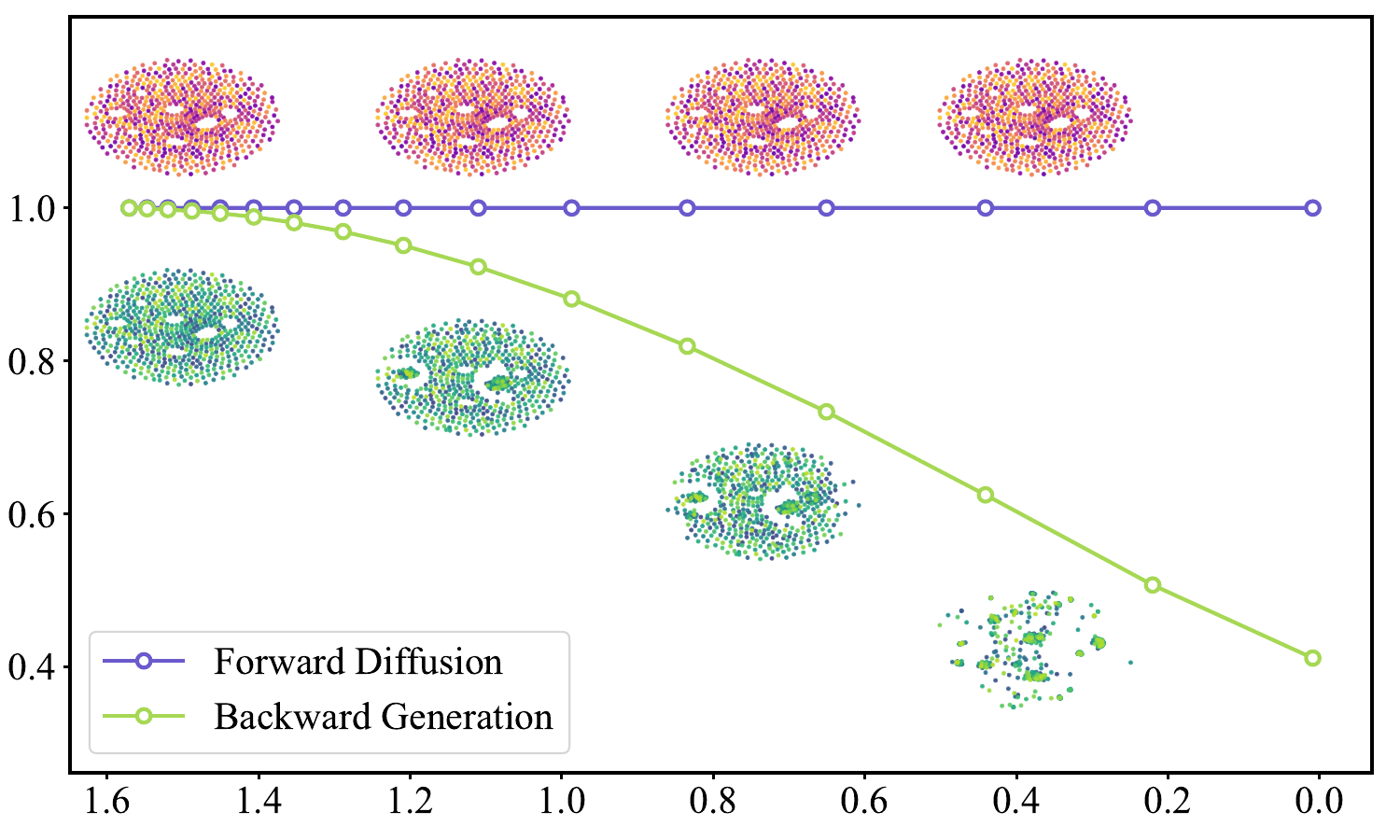}
    \caption{Evolution of Equivalent Noise Distributions.
    \textbf{\textit{Curves:}} Cosine similarity between timestep-specific equivalent noise and initial noise.
    \textbf{\textit{Scatter Plots:}} T-SNE projections of equivalent noise distributions at selected timesteps.}
    \vspace{-0.5em}
    \label{supfig:VisDis}
\end{figure}

As discussed in Sec.~\ref{sec:tbcm}, to verify that the differences in equivalent noise between the forward and backward processes are distributional rather than instance-specific, we visualize the overall distribution of equivalent noise during the diffusion and generation processes using t-SNE, as shown in Fig.~\ref{supfig:VisDis}. 

The results show that in the diffusion process, the distribution of equivalent noise remains consistent, reflecting the instance-level consistency shown in Fig.~\ref{fig:for-back}. In contrast, during the generation process, the noise distribution exhibits significant pattern changes, and its similarity to the initial noise gradually decreases throughout the process.

\section{Comparison of Sampling Strategies.}
\label{sup:sampling}
We compare the predictions of $\bx_0$ by the teacher model under different sampling schemes and inference steps, based on the hypothesis that better $\bx_0$ predictions partially reflect higher overall sample quality along the entire trajectory. 

\begin{figure}[h]
    \centering
    \subfloat[Visualizations of predicted $\mathbf{x}_0$ under different sampling schemes.]
    {
        \includegraphics[width=\linewidth]{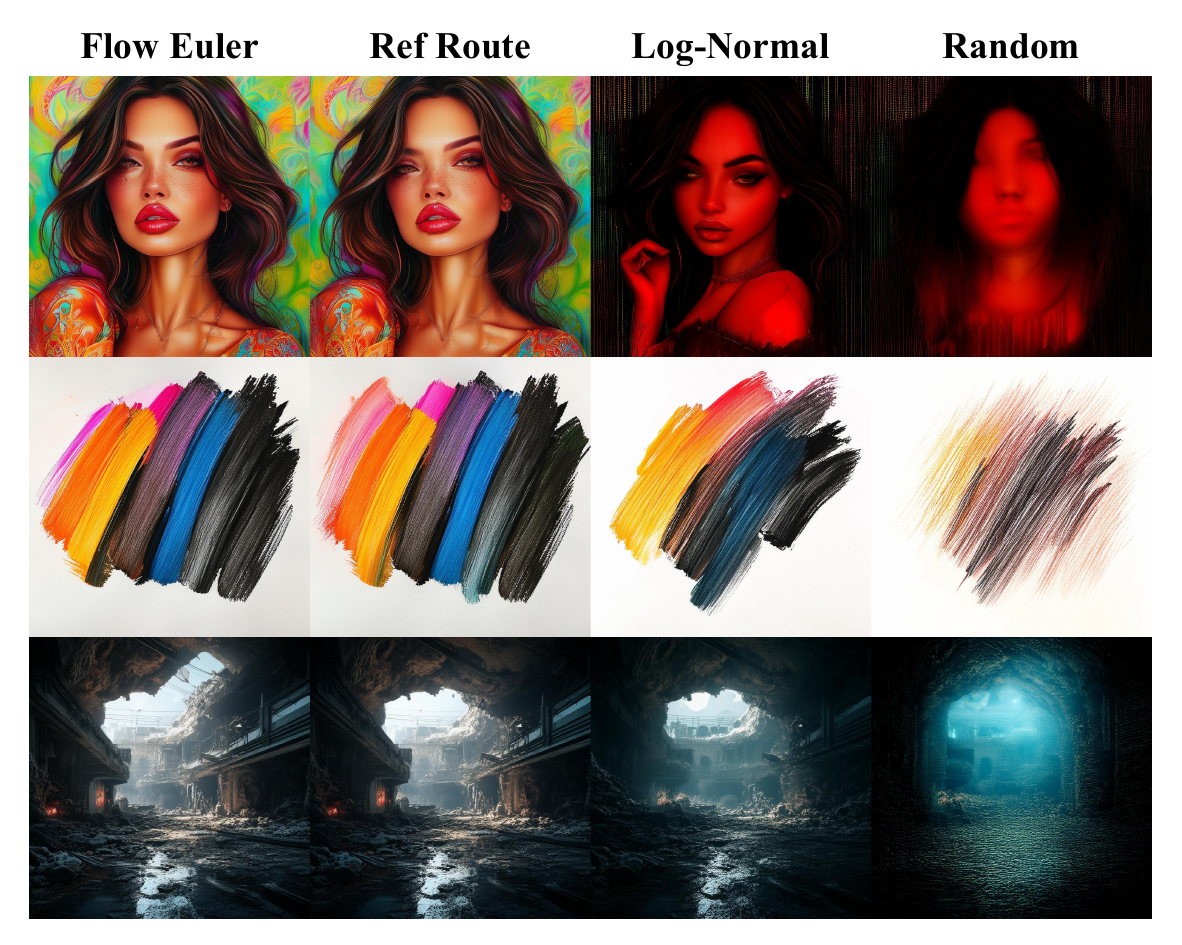}
        \label{supfig:schemes}
    } \\[-2pt]
    \subfloat[Visualizations of predicted $\mathbf{x}_0$ under different sampling steps.]
     {
        \includegraphics[width=\linewidth]{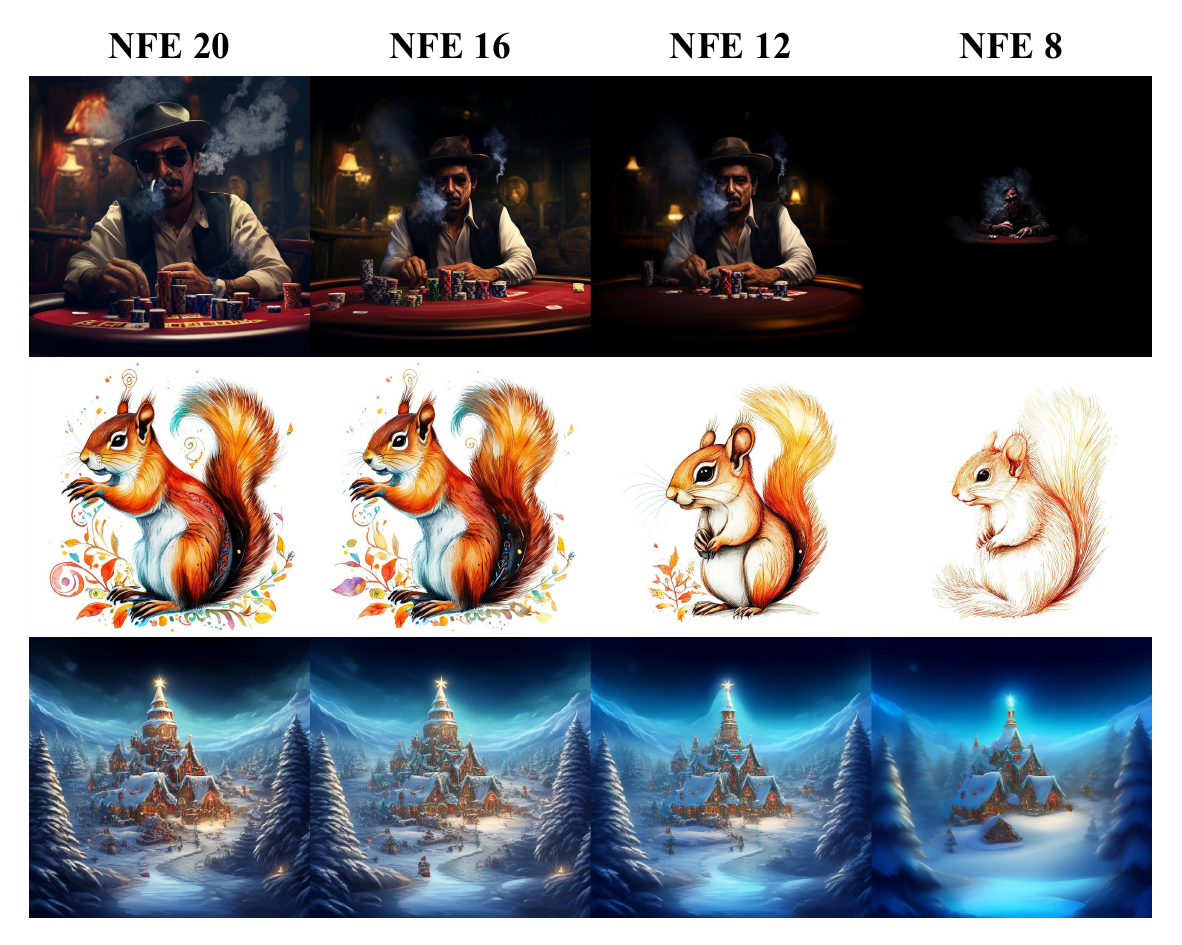}
        \label{supfig:steps}
    }
    \caption{Comparison of predicted $\bx_0$ across (a) different sampling schemes and (b) different sampling steps.}
    \label{supfig:sampling}
\end{figure}

As shown in Fig.~\ref{supfig:sampling}, the Reference Route sampling scheme consistently produces high-quality $\bx_0$ predictions close to those obtained with the Flow Euler Scheduler, followed by the Logit-Normal scheme, and then Random sampling, which aligns with the experimental results in Tab.~\ref{table_ablation_schemes}. On the other hand, increasing the number of inference steps generally improves the quality of predicted $\bx_0$, but the differences gradually diminish as the number of steps increases, consistent with the observations reported in Tab.~\ref{table_ablation_steps}.

\section{Algorithm of TBCM.}
\label{sup:algorithm}
We provide the algorithm for the TBCM paradigm, with key differences from sCM highlighted in blue. As shown in ~\cref{alg:TBCM}, TBCM does not require any image data and collects training samples along the teacher's inference trajectory, which are then used to compute the sCM loss. The algorithm also incorporates the brightness filter and stability hyperparameter adjustment introduced in Sec.~\ref{sec:additional}.

\section{Illustration of the Brightness Filter.}
\label{sup:brightness}
\begin{figure}[h]
    \centering
    \includegraphics[width=0.5\textwidth]{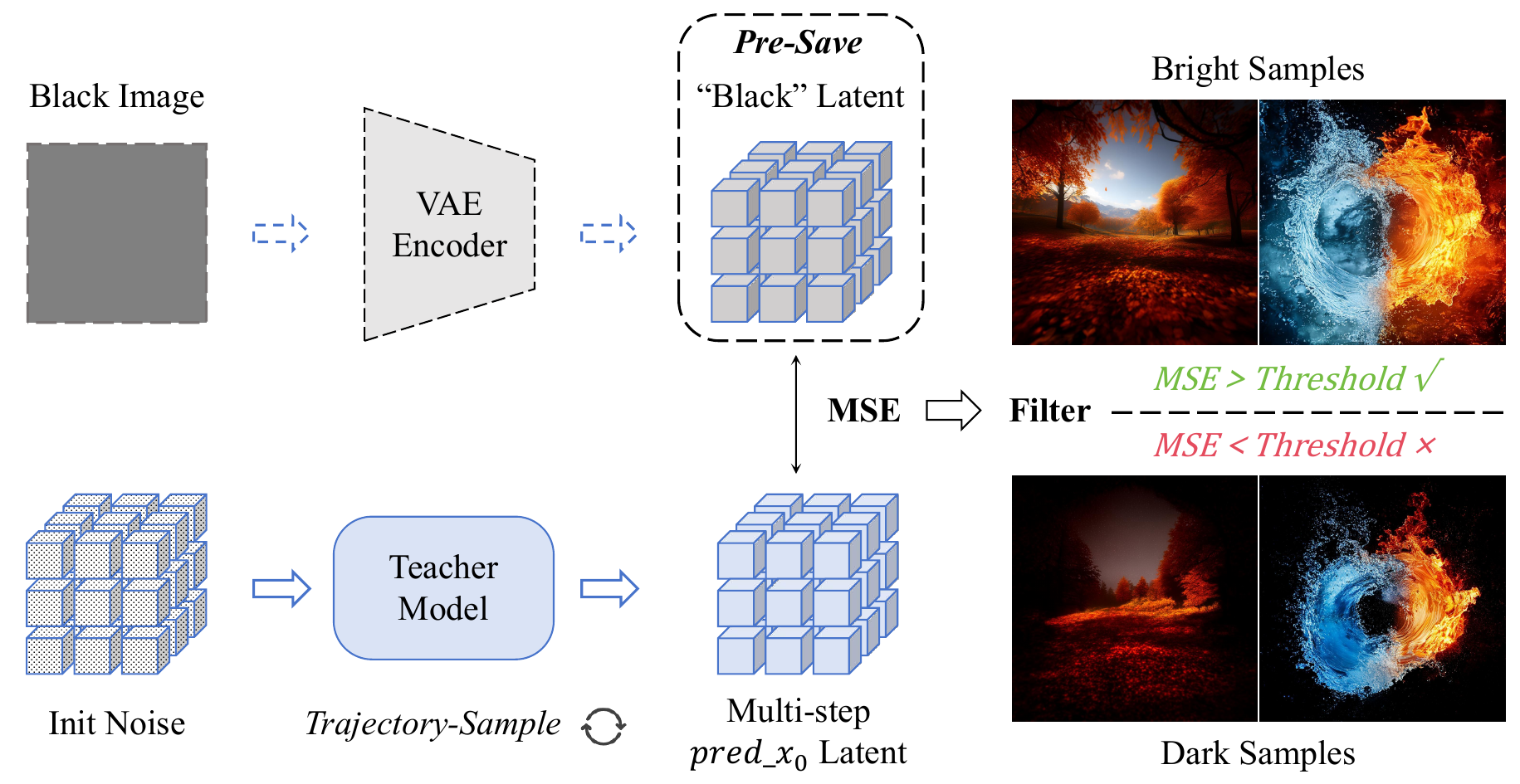}
    \caption{Illustration of the Brightness Filter Strategy. Low-quality generated latent samples, often unusually dark, are identified by measuring their similarity to a completely black latent representation and filtered using a simple threshold.}
    \label{supfig:Brightness}
\end{figure}

We provide an illustration of the Brightness Filter strategy mentioned in Sec.~\ref{sec:additional}, as shown in Fig.~\ref{supfig:Brightness}. This strategy aims to directly identify low-quality samples in the latent space without decoding them to pixel space via the VAE. Since low-quality samples generated by the teacher model are often observed to be unusually dark, this issue can be addressed by filtering out samples with low brightness directly in the latent space. To this end, we precompute the latent representation of a completely black image and measure its similarity to latent-space samples, followed by a simple threshold-based filtering.

\section{Multi-Step Generation Results.}
\label{sup:generation}
\begin{figure*}[th]
    \centering
    \includegraphics[width=1.0\textwidth]{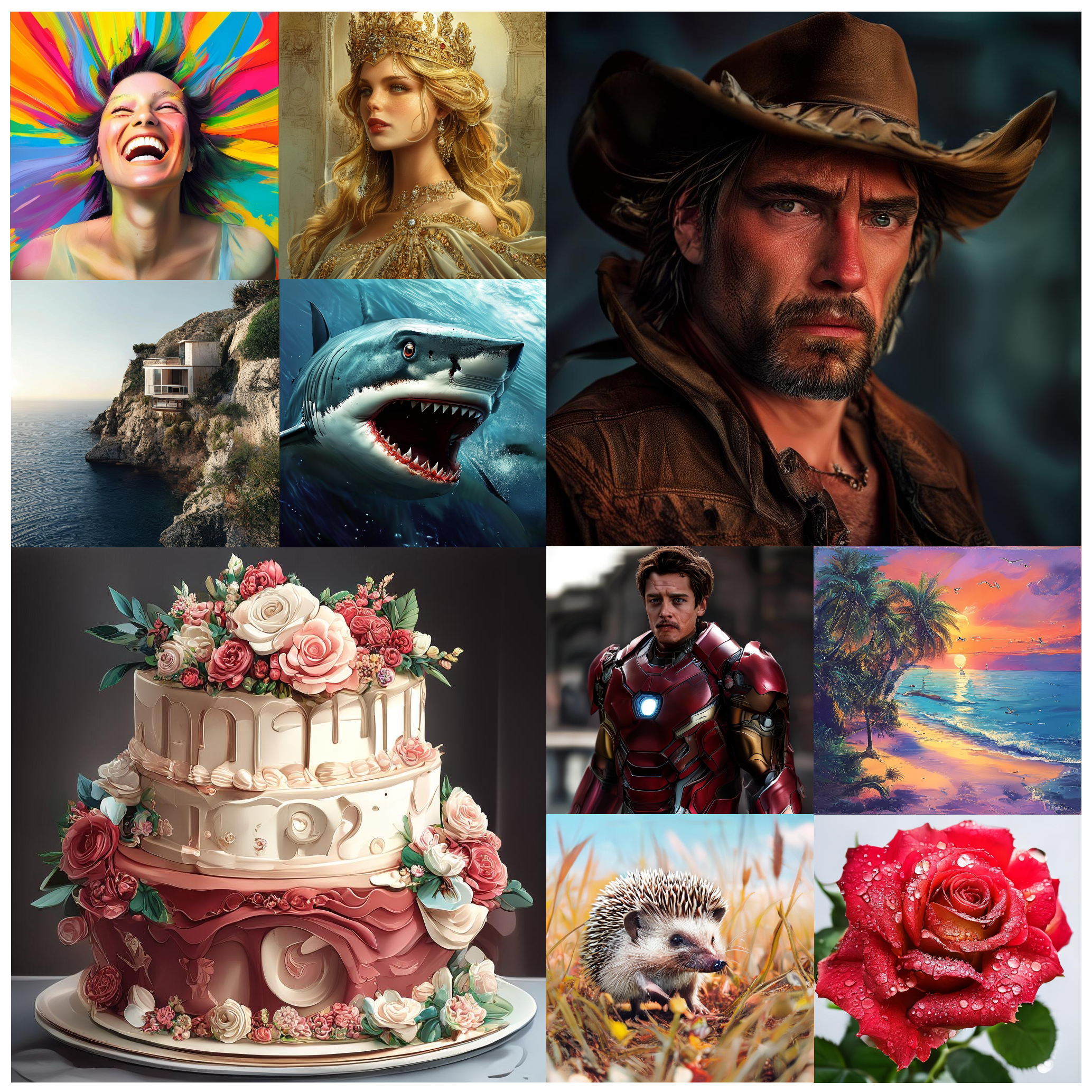}
    \caption{Two Step Generation Results.}
    \label{supfig:generate n2}
\end{figure*}

\begin{figure*}[th]
    \centering
    \includegraphics[width=1.0\textwidth]{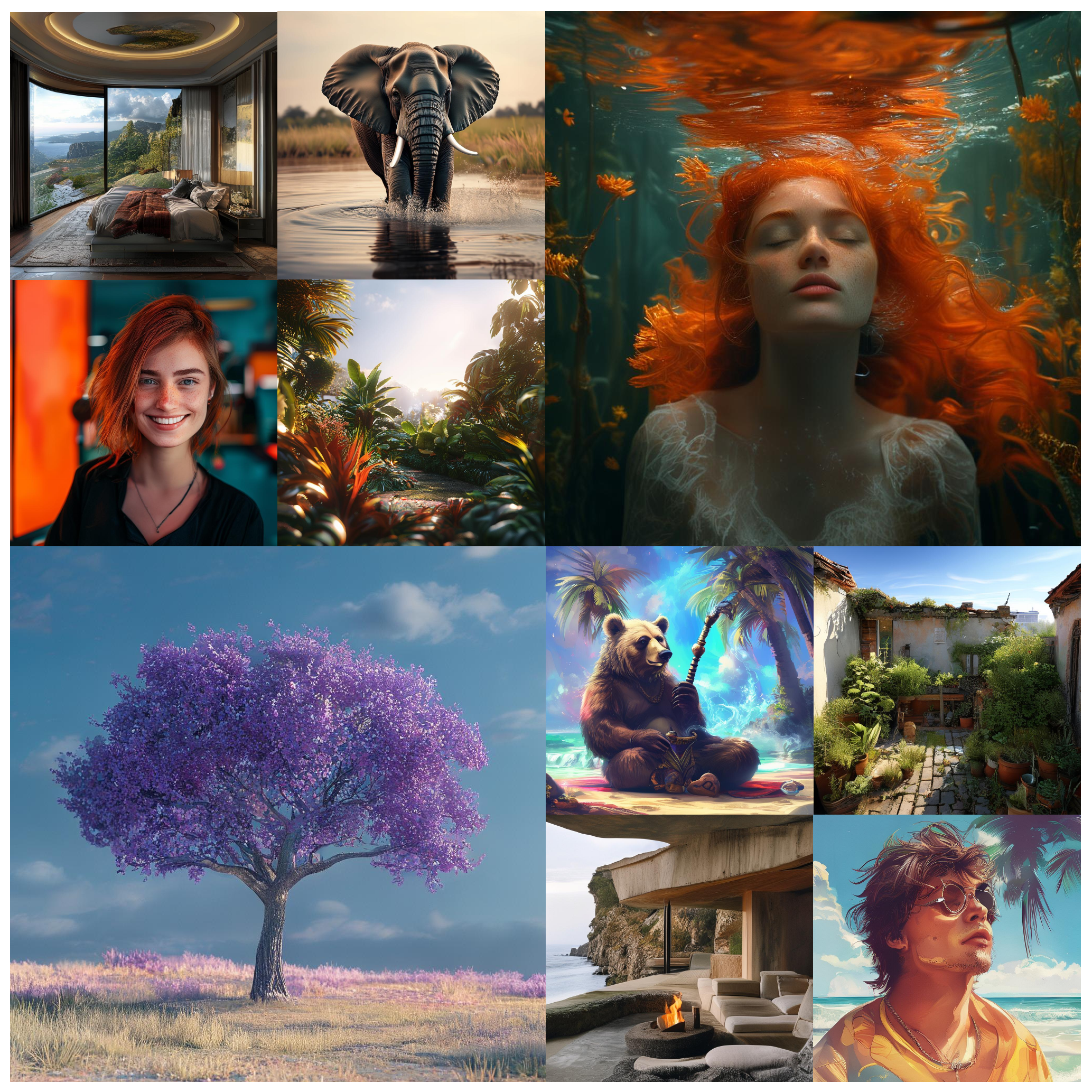}
    \caption{Four Step Generation Results.}
    \label{supfig:generate n4}
\end{figure*}

Using the scheduler described in CMs~\citep{cm}, which first maps back to $\bx_0$ and then adds noise to an intermediate timestep, our method can also generate images at different inference steps. Here, we provide results for 2-step (Fig.~\ref{supfig:generate n2}) and 4-step (Fig.~\ref{supfig:generate n4}) inference, while the 1-step results (Fig.~\ref{fig:generate}) are presented in the main paper.

%% file: main.bbl
\begin{thebibliography}{47}
\providecommand{\natexlab}[1]{#1}
\providecommand{\url}[1]{\texttt{#1}}
\expandafter\ifx\csname urlstyle\endcsname\relax
  \providecommand{\doi}[1]{doi: #1}\else
  \providecommand{\doi}{doi: \begingroup \urlstyle{rm}\Url}\fi

\bibitem[Albergo and Vanden-Eijnden()]{albergobuilding}
Michael~Samuel Albergo and Eric Vanden-Eijnden.
\newblock Building normalizing flows with stochastic interpolants.
\newblock In \emph{The Eleventh International Conference on Learning Representations}.

\bibitem[Batifol et~al.(2025)Batifol, Blattmann, Boesel, Consul, Diagne, Dockhorn, English, English, Esser, Kulal, et~al.]{flux}
Stephen Batifol, Andreas Blattmann, Frederic Boesel, Saksham Consul, Cyril Diagne, Tim Dockhorn, Jack English, Zion English, Patrick Esser, Sumith Kulal, et~al.
\newblock Flux. 1 kontext: Flow matching for in-context image generation and editing in latent space.
\newblock \emph{arXiv e-prints}, pages arXiv--2506, 2025.

\bibitem[Blattmann et~al.(2023)Blattmann, Dockhorn, Kulal, Mendelevitch, Kilian, Lorenz, Levi, English, Voleti, Letts, et~al.]{svd}
Andreas Blattmann, Tim Dockhorn, Sumith Kulal, Daniel Mendelevitch, Maciej Kilian, Dominik Lorenz, Yam Levi, Zion English, Vikram Voleti, Adam Letts, et~al.
\newblock Stable video diffusion: Scaling latent video diffusion models to large datasets.
\newblock \emph{arXiv preprint arXiv:2311.15127}, 2023.

\bibitem[Chen et~al.(2023)Chen, Xia, He, Zhang, Cun, Yang, Xing, Liu, Chen, Wang, et~al.]{videocrafter}
Haoxin Chen, Menghan Xia, Yingqing He, Yong Zhang, Xiaodong Cun, Shaoshu Yang, Jinbo Xing, Yaofang Liu, Qifeng Chen, Xintao Wang, et~al.
\newblock Videocrafter1: Open diffusion models for high-quality video generation.
\newblock \emph{arXiv preprint arXiv:2310.19512}, 2023.

\bibitem[Chen et~al.({\natexlab{a}})Chen, Cai, Chen, Xie, Yang, Tang, Li, and Han]{dcae}
Junyu Chen, Han Cai, Junsong Chen, Enze Xie, Shang Yang, Haotian Tang, Muyang Li, and Song Han.
\newblock Deep compression autoencoder for efficient high-resolution diffusion models.
\newblock In \emph{The Thirteenth International Conference on Learning Representations}, {\natexlab{a}}.

\bibitem[Chen et~al.({\natexlab{b}})Chen, Luo, and Xie]{pixart-delta}
Junsong Chen, Simian Luo, and Enze Xie.
\newblock Pixart-$\delta$: Fast and controllable image generation with latent consistency models.
\newblock In \emph{ICML 2024 Workshop on Theoretical Foundations of Foundation Models}, {\natexlab{b}}.

\bibitem[Chen et~al.(2024{\natexlab{a}})Chen, Ge, Xie, Wu, Yao, Ren, Wang, Luo, Lu, and Li]{pixart-sigma}
Junsong Chen, Chongjian Ge, Enze Xie, Yue Wu, Lewei Yao, Xiaozhe Ren, Zhongdao Wang, Ping Luo, Huchuan Lu, and Zhenguo Li.
\newblock Pixart-$\sigma$: Weak-to-strong training of diffusion transformer for 4k text-to-image generation.
\newblock In \emph{European Conference on Computer Vision}, pages 74--91. Springer, 2024{\natexlab{a}}.

\bibitem[Chen et~al.(2024{\natexlab{b}})Chen, Yu, Ge, Yao, Xie, Wang, Kwok, Luo, Lu, and Li]{pixart}
Junsong Chen, Jincheng Yu, Chongjian Ge, Lewei Yao, Enze Xie, Zhongdao Wang, James~T Kwok, Ping Luo, Huchuan Lu, and Zhenguo Li.
\newblock Pixart-$\alpha$: Fast training of diffusion transformer for photorealistic text-to-image synthesis.
\newblock In \emph{ICLR}, 2024{\natexlab{b}}.

\bibitem[Chen et~al.(2025)Chen, Xue, Zhao, Yu, Paul, Chen, Cai, Han, and Xie]{sana-sprint}
Junsong Chen, Shuchen Xue, Yuyang Zhao, Jincheng Yu, Sayak Paul, Junyu Chen, Han Cai, Song Han, and Enze Xie.
\newblock Sana-sprint: One-step diffusion with continuous-time consistency distillation.
\newblock \emph{arXiv preprint arXiv:2503.09641}, 2025.

\bibitem[Chen et~al.(2024{\natexlab{c}})Chen, Shen, Ye, Cao, Tu, Bouganis, Zhao, and Chen]{delta-dit}
Pengtao Chen, Mingzhu Shen, Peng Ye, Jianjian Cao, Chongjun Tu, Christos-Savvas Bouganis, Yiren Zhao, and Tao Chen.
\newblock $\delta$-dit: A training-free acceleration method tailored for diffusion transformers.
\newblock \emph{ArXiv}, abs/2406.01125, 2024{\natexlab{c}}.

\bibitem[Esser et~al.(2024)Esser, Kulal, Blattmann, Entezari, M{\"u}ller, Saini, Levi, Lorenz, Sauer, Boesel, et~al.]{sd3}
Patrick Esser, Sumith Kulal, Andreas Blattmann, Rahim Entezari, Jonas M{\"u}ller, Harry Saini, Yam Levi, Dominik Lorenz, Axel Sauer, Frederic Boesel, et~al.
\newblock Scaling rectified flow transformers for high-resolution image synthesis.
\newblock In \emph{Proceedings of the 41st International Conference on Machine Learning}, pages 12606--12633, 2024.

\bibitem[Geng et~al.()Geng, Pokle, Luo, Lin, and Kolter]{ect}
Zhengyang Geng, Ashwini Pokle, Weijian Luo, Justin Lin, and J~Zico Kolter.
\newblock Consistency models made easy.
\newblock In \emph{The Thirteenth International Conference on Learning Representations}.

\bibitem[Heek et~al.(2024)Heek, Hoogeboom, and Salimans]{mcm}
Jonathan Heek, Emiel Hoogeboom, and Tim Salimans.
\newblock Multistep consistency models.
\newblock \emph{arXiv preprint arXiv:2403.06807}, 2024.

\bibitem[Hertz et~al.()Hertz, Mokady, Tenenbaum, Aberman, Pritch, and Cohen-or]{prompt-to-prompt}
Amir Hertz, Ron Mokady, Jay Tenenbaum, Kfir Aberman, Yael Pritch, and Daniel Cohen-or.
\newblock Prompt-to-prompt image editing with cross-attention control.
\newblock In \emph{The Eleventh International Conference on Learning Representations}.

\bibitem[Ho et~al.(2020)Ho, Jain, and Abbeel]{ddpm}
Jonathan Ho, Ajay Jain, and Pieter Abbeel.
\newblock Denoising diffusion probabilistic models.
\newblock \emph{Advances in neural information processing systems}, 33:\penalty0 6840--6851, 2020.

\bibitem[Karras et~al.(2022)Karras, Aittala, Aila, and Laine]{edm}
Tero Karras, Miika Aittala, Timo Aila, and Samuli Laine.
\newblock Elucidating the design space of diffusion-based generative models.
\newblock \emph{Advances in neural information processing systems}, 35:\penalty0 26565--26577, 2022.

\bibitem[Karras et~al.(2024)Karras, Aittala, Lehtinen, Hellsten, Aila, and Laine]{edm2}
Tero Karras, Miika Aittala, Jaakko Lehtinen, Janne Hellsten, Timo Aila, and Samuli Laine.
\newblock Analyzing and improving the training dynamics of diffusion models.
\newblock In \emph{Proceedings of the IEEE/CVF Conference on Computer Vision and Pattern Recognition}, pages 24174--24184, 2024.

\bibitem[Kim et~al.()Kim, Lai, Liao, Murata, Takida, Uesaka, He, Mitsufuji, and Ermon]{ctm}
Dongjun Kim, Chieh-Hsin Lai, Wei-Hsiang Liao, Naoki Murata, Yuhta Takida, Toshimitsu Uesaka, Yutong He, Yuki Mitsufuji, and Stefano Ermon.
\newblock Consistency trajectory models: Learning probability flow ode trajectory of diffusion.
\newblock In \emph{The Twelfth International Conference on Learning Representations}.

\bibitem[Li et~al.(2024)Li, Kamko, Akhgari, Sabet, Xu, and Doshi]{playground}
Daiqing Li, Aleks Kamko, Ehsan Akhgari, Ali Sabet, Linmiao Xu, and Suhail Doshi.
\newblock Playground v2. 5: Three insights towards enhancing aesthetic quality in text-to-image generation.
\newblock \emph{arXiv preprint arXiv:2402.17245}, 2024.

\bibitem[Lin et~al.(2024)Lin, Ge, Cheng, Li, Zhu, Wang, He, Ye, Yuan, Chen, et~al.]{opensora_plan}
Bin Lin, Yunyang Ge, Xinhua Cheng, Zongjian Li, Bin Zhu, Shaodong Wang, Xianyi He, Yang Ye, Shenghai Yuan, Liuhan Chen, et~al.
\newblock Open-sora plan: Open-source large video generation model.
\newblock \emph{arXiv preprint arXiv:2412.00131}, 2024.

\bibitem[Lipman et~al.(2023)Lipman, Chen, Ben-Hamu, Nickel, and Le]{flow-matching}
Yaron Lipman, Ricky~TQ Chen, Heli Ben-Hamu, Maximilian Nickel, and Matt Le.
\newblock Flow matching for generative modeling.
\newblock In \emph{11th International Conference on Learning Representations, ICLR 2023}, 2023.

\bibitem[Liu et~al.(2023)Liu, Ning, Lin, Yang, and Wang]{oms-dpm}
Enshu Liu, Xuefei Ning, Zinan Lin, Huazhong Yang, and Yu Wang.
\newblock Oms-dpm: Optimizing the model schedule for diffusion probabilistic models.
\newblock In \emph{International Conference on Machine Learning}, pages 21915--21936. PMLR, 2023.

\bibitem[Liu et~al.()Liu, Gong, et~al.]{flow-straight}
Xingchao Liu, Chengyue Gong, et~al.
\newblock Flow straight and fast: Learning to generate and transfer data with rectified flow.
\newblock In \emph{The Eleventh International Conference on Learning Representations}.

\bibitem[Lu and Song()]{scm}
Cheng Lu and Yang Song.
\newblock Simplifying, stabilizing and scaling continuous-time consistency models.
\newblock In \emph{The Thirteenth International Conference on Learning Representations}.

\bibitem[Lu et~al.(2022)Lu, Zhou, Bao, Chen, Li, and Zhu]{dpm-solver}
Cheng Lu, Yuhao Zhou, Fan Bao, Jianfei Chen, Chongxuan Li, and Jun Zhu.
\newblock Dpm-solver: A fast ode solver for diffusion probabilistic model sampling in around 10 steps.
\newblock \emph{Advances in neural information processing systems}, 35:\penalty0 5775--5787, 2022.

\bibitem[Luhman and Luhman(2021)]{direct}
Eric Luhman and Troy Luhman.
\newblock Knowledge distillation in iterative generative models for improved sampling speed.
\newblock \emph{arXiv preprint arXiv:2101.02388}, 2021.

\bibitem[Luo et~al.(2023)Luo, Tan, Huang, Li, and Zhao]{lcm}
Simian Luo, Yiqin Tan, Longbo Huang, Jian Li, and Hang Zhao.
\newblock Latent consistency models: Synthesizing high-resolution images with few-step inference.
\newblock \emph{arXiv preprint arXiv:2310.04378}, 2023.

\bibitem[Luo et~al.(2024)Luo, Huang, Geng, Kolter, and Qi]{sim}
Weijian Luo, Zemin Huang, Zhengyang Geng, J~Zico Kolter, and Guo-jun Qi.
\newblock One-step diffusion distillation through score implicit matching.
\newblock \emph{Advances in Neural Information Processing Systems}, 37:\penalty0 115377--115408, 2024.

\bibitem[Nichol and Dhariwal(2021)]{iddpm}
Alexander~Quinn Nichol and Prafulla Dhariwal.
\newblock Improved denoising diffusion probabilistic models.
\newblock In \emph{International conference on machine learning}, pages 8162--8171. PMLR, 2021.

\bibitem[Podell et~al.()Podell, English, Lacey, Blattmann, Dockhorn, M{\"u}ller, Penna, and Rombach]{sdxl}
Dustin Podell, Zion English, Kyle Lacey, Andreas Blattmann, Tim Dockhorn, Jonas M{\"u}ller, Joe Penna, and Robin Rombach.
\newblock Sdxl: Improving latent diffusion models for high-resolution image synthesis.
\newblock In \emph{The Twelfth International Conference on Learning Representations}.

\bibitem[Poole et~al.()Poole, Jain, Barron, and Mildenhall]{dreamfusion}
Ben Poole, Ajay Jain, Jonathan~T Barron, and Ben Mildenhall.
\newblock Dreamfusion: Text-to-3d using 2d diffusion.
\newblock In \emph{The Eleventh International Conference on Learning Representations}.

\bibitem[Ramesh et~al.(2021)Ramesh, Pavlov, Goh, Gray, Voss, Radford, Chen, and Sutskever]{dalle}
Aditya Ramesh, Mikhail Pavlov, Gabriel Goh, Scott Gray, Chelsea Voss, Alec Radford, Mark Chen, and Ilya Sutskever.
\newblock Zero-shot text-to-image generation.
\newblock In \emph{International conference on machine learning}, pages 8821--8831. Pmlr, 2021.

\bibitem[Rombach et~al.(2022)Rombach, Blattmann, Lorenz, Esser, and Ommer]{ldm}
Robin Rombach, Andreas Blattmann, Dominik Lorenz, Patrick Esser, and Bj{\"o}rn Ommer.
\newblock High-resolution image synthesis with latent diffusion models.
\newblock In \emph{Proceedings of the IEEE/CVF conference on computer vision and pattern recognition}, pages 10684--10695, 2022.

\bibitem[Saharia et~al.(2022)Saharia, Chan, Saxena, Li, Whang, Denton, Ghasemipour, Gontijo~Lopes, Karagol~Ayan, Salimans, et~al.]{imagen}
Chitwan Saharia, William Chan, Saurabh Saxena, Lala Li, Jay Whang, Emily~L Denton, Kamyar Ghasemipour, Raphael Gontijo~Lopes, Burcu Karagol~Ayan, Tim Salimans, et~al.
\newblock Photorealistic text-to-image diffusion models with deep language understanding.
\newblock \emph{Advances in neural information processing systems}, 35:\penalty0 36479--36494, 2022.

\bibitem[Salimans and Ho()]{progressive}
Tim Salimans and Jonathan Ho.
\newblock Progressive distillation for fast sampling of diffusion models.
\newblock In \emph{International Conference on Learning Representations}.

\bibitem[Sauer et~al.(2024{\natexlab{a}})Sauer, Boesel, Dockhorn, Blattmann, Esser, and Rombach]{ladd}
Axel Sauer, Frederic Boesel, Tim Dockhorn, Andreas Blattmann, Patrick Esser, and Robin Rombach.
\newblock Fast high-resolution image synthesis with latent adversarial diffusion distillation.
\newblock In \emph{SIGGRAPH Asia 2024 Conference Papers}, pages 1--11, 2024{\natexlab{a}}.

\bibitem[Sauer et~al.(2024{\natexlab{b}})Sauer, Lorenz, Blattmann, and Rombach]{add}
Axel Sauer, Dominik Lorenz, Andreas Blattmann, and Robin Rombach.
\newblock Adversarial diffusion distillation.
\newblock In \emph{European Conference on Computer Vision}, pages 87--103. Springer, 2024{\natexlab{b}}.

\bibitem[Song et~al.()Song, Meng, and Ermon]{ddim}
Jiaming Song, Chenlin Meng, and Stefano Ermon.
\newblock Denoising diffusion implicit models.
\newblock In \emph{International Conference on Learning Representations}.

\bibitem[Song et~al.(2023)Song, Dhariwal, Chen, and Sutskever]{cm}
Yang Song, Prafulla Dhariwal, Mark Chen, and Ilya Sutskever.
\newblock Consistency models.
\newblock In \emph{Proceedings of the 40th International Conference on Machine Learning}, pages 32211--32252, 2023.

\bibitem[Wan et~al.(2025)Wan, Wang, Ai, Wen, Mao, Xie, Chen, Yu, Zhao, Yang, et~al.]{wan}
Team Wan, Ang Wang, Baole Ai, Bin Wen, Chaojie Mao, Chen-Wei Xie, Di Chen, Feiwu Yu, Haiming Zhao, Jianxiao Yang, et~al.
\newblock Wan: Open and advanced large-scale video generative models.
\newblock \emph{arXiv preprint arXiv:2503.20314}, 2025.

\bibitem[Wang et~al.(2024)Wang, Huang, Bergman, Shen, Gao, Lingelbach, Sun, Bian, Song, Liu, et~al.]{pcm}
Fu-Yun Wang, Zhaoyang Huang, Alexander Bergman, Dazhong Shen, Peng Gao, Michael Lingelbach, Keqiang Sun, Weikang Bian, Guanglu Song, Yu Liu, et~al.
\newblock Phased consistency models.
\newblock \emph{Advances in neural information processing systems}, 37:\penalty0 83951--84009, 2024.

\bibitem[Wang et~al.(2023)Wang, Lu, Wang, Bao, Li, Su, and Zhu]{prolificdreamer}
Zhengyi Wang, Cheng Lu, Yikai Wang, Fan Bao, Chongxuan Li, Hang Su, and Jun Zhu.
\newblock Prolificdreamer: High-fidelity and diverse text-to-3d generation with variational score distillation.
\newblock \emph{Advances in neural information processing systems}, 36:\penalty0 8406--8441, 2023.

\bibitem[Xie et~al.(2024)Xie, Chen, Chen, Cai, Tang, Lin, Zhang, Li, Zhu, Lu, et~al.]{sana}
Enze Xie, Junsong Chen, Junyu Chen, Han Cai, Haotian Tang, Yujun Lin, Zhekai Zhang, Muyang Li, Ligeng Zhu, Yao Lu, et~al.
\newblock Sana: Efficient high-resolution image synthesis with linear diffusion transformers.
\newblock \emph{arXiv preprint arXiv:2410.10629}, 2024.

\bibitem[Yang et~al.()Yang, Teng, Zheng, Ding, Huang, Xu, Yang, Hong, Zhang, Feng, et~al.]{cogvideox}
Zhuoyi Yang, Jiayan Teng, Wendi Zheng, Ming Ding, Shiyu Huang, Jiazheng Xu, Yuanming Yang, Wenyi Hong, Xiaohan Zhang, Guanyu Feng, et~al.
\newblock Cogvideox: Text-to-video diffusion models with an expert transformer.
\newblock In \emph{The Thirteenth International Conference on Learning Representations}.

\bibitem[Yin et~al.(2024{\natexlab{a}})Yin, Gharbi, Park, Zhang, Shechtman, Durand, and Freeman]{dmd2}
Tianwei Yin, Micha{\"e}l Gharbi, Taesung Park, Richard Zhang, Eli Shechtman, Fredo Durand, and Bill Freeman.
\newblock Improved distribution matching distillation for fast image synthesis.
\newblock \emph{Advances in neural information processing systems}, 37:\penalty0 47455--47487, 2024{\natexlab{a}}.

\bibitem[Yin et~al.(2024{\natexlab{b}})Yin, Gharbi, Zhang, Shechtman, Durand, Freeman, and Park]{dmd}
Tianwei Yin, Micha{\"e}l Gharbi, Richard Zhang, Eli Shechtman, Fredo Durand, William~T Freeman, and Taesung Park.
\newblock One-step diffusion with distribution matching distillation.
\newblock In \emph{Proceedings of the IEEE/CVF conference on computer vision and pattern recognition}, pages 6613--6623, 2024{\natexlab{b}}.

\bibitem[Zhou et~al.(2024)Zhou, Zheng, Wang, Yin, and Huang]{sid}
Mingyuan Zhou, Huangjie Zheng, Zhendong Wang, Mingzhang Yin, and Hai Huang.
\newblock Score identity distillation: Exponentially fast distillation of pretrained diffusion models for one-step generation.
\newblock In \emph{Forty-first International Conference on Machine Learning}, 2024.

\end{thebibliography}
